\documentclass[lettersize,journal]{IEEEtran}
\usepackage{amsmath,amsfonts}
\usepackage{algorithmic}
\usepackage{algorithm}
\usepackage{array}
\usepackage[caption=false,font=normalsize,labelfont=sf,textfont=sf]{subfig}
\usepackage{textcomp}
\usepackage{stfloats}
\usepackage{url}
\usepackage{verbatim}
\usepackage{graphicx}
\usepackage{cite}
\usepackage{amsmath}
\usepackage{amssymb}
\usepackage{xcolor}
\usepackage{graphicx}
\usepackage{caption}
\usepackage{hyperref}
\usepackage{booktabs}
\usepackage{cleveref}

\usepackage{multirow}

\usepackage{algorithm}
\usepackage{algorithmic}
\usepackage{subcaption}
\usepackage{cite}
\usepackage{mathtools}

\newtheorem{assumption}{Assumption}

\newtheorem{lemma}{Lemma}

\newtheorem{theorem}{Theorem}

\newcommand{\mc}{\mathcal}
\newcommand{\mb}{\mathbb}
\newcommand{\mf}{\mathbf}

\DeclareMathOperator*{\argmax}{argmax}
\DeclareMathOperator*{\argmin}{argmin}

\newcommand{\qedbox}{\hfill$\square$}
\newenvironment{proofof}[1]{%
  \par\noindent\textit{Proof of #1:}\ %
}{\qedbox\par}

\begin{document}

\title{Contraction-Aware Reinforcement Learning for Nonlinear Control with Statistical Robustness}


\author{Minjae Cho, Hiroyasu Tsukamoto, and Huy T. Tran
\thanks{The authors are with Department of Aerospace Engineering, The Grainger College of Engineering, University of Illinois Urbana-Champaign, Urbana, IL 61801 USA. (e-mail: \{minjae5, hiroyasu, huytran1\}@illinois.edu)
}}

\markboth{Journal of \LaTeX\ Class Files,~Vol.~14, No.~8, August~2021}%
{Shell \MakeLowercase{\textit{et al.}}: A Sample Article Using IEEEtran.cls for IEEE Journals}


\maketitle

\begin{abstract}
    Control contraction metrics (CCMs)---defined by Riemannian metrics under which a closed-loop system is incrementally exponentially stable---offer a constructive framework for synthesizing contracting policies in nonlinear path-tracking problems.
    However, while the synthesized policies ensure pointwise satisfaction of the CCM conditions, they may not ensure long-term optimality (i.e., minimizing cumulative trajectory-level tracking error) over both transient and steady-state regimes. Furthermore, the myopic nature of these policies could also make them more susceptible to learning biases when approximate dynamics are used to formulate CCMs.
    To address these issues, we propose to integrate CCMs into reinforcement learning (RL).
    CCMs provide dynamics-informed feedback for learning a policy that has a stability guarantee---i.e., is contraction-aware---while RL provides a framework for minimizing cumulative tracking error under approximate dynamics.
    Given a pretrained dynamics model, our algorithm, contraction-aware RL (CARL), simultaneously learns to generate CCMs and optimize a policy for rewards defined by those CCMs.
    We demonstrate that CARL enhances path-tracking performance and is robust to errors in approximated dynamics compared to relevant baselines in both simulated and real-world robot experiments.
    We also provide theoretical rationale for integrating CCMs into RL.
    Our code is available at \url{https://github.com/Mgineer117/CARL},
    and a video of our real-world robot experiments can be found at \url{https://youtu.be/sOJ4hulbop0}.

\end{abstract}


\section{Introduction}\label{sec:introduction}

A fundamental challenge in designing path-tracking control policies, particularly in safety-critical nonlinear systems, is to ensure their reliable operation while efficiently achieving task objectives.
Control contraction metrics (CCMs) \cite{manchester2017control}, grounded in contraction theory \cite{LOHMILLER1998683}, provide necessary and sufficient conditions for certifying incremental and exponential stabilizability through a path-tracking policy (i.e., a contracting policy) with respect to an arbitrary feasible reference trajectory.
This certificate is contingent upon the existence of CCMs, Riemannian metrics under which the closed-loop system is guaranteed to achieve contraction.
The ability to provide such a certificate has motivated various methods to synthesize CCMs and corresponding contracting policies for several classes of nonlinear systems.

One approach first identifies CCMs by solving an infinite-dimensional convex feasibility problem using techniques such as sum-of-squares (SoS) programming~\cite{manchester2017control,singh2017robust}, finite difference methods
\cite{tsukamoto2020robust,tsukamoto2020neural}, or reproducing kernel Hilbert space theory for partially known dynamics \cite{singh_learning_2018}. These metrics can be optimized further, e.g., by minimizing an upper bound on the steady-state tracking error subject to the CCM constraints~\cite{singh2017robust,tsukamoto2020robust}. 
The identified CCMs can then be used to synthesize a differential feedback control policy, whose path-integrated form guarantees contraction.
However, these methods scale poorly to high-dimensional nonlinear systems as they require extensive computation to derive controls (e.g., solving the feasibility problem for CCMs and finding minimizing geodesics defined by CCMs) \cite{dawson2023safe}. They may also rely on restrictive assumptions on the structure of the system dynamics (e.g., polynomial dynamics).

\begin{figure}
    \centering
    \includegraphics[width=0.85\linewidth]{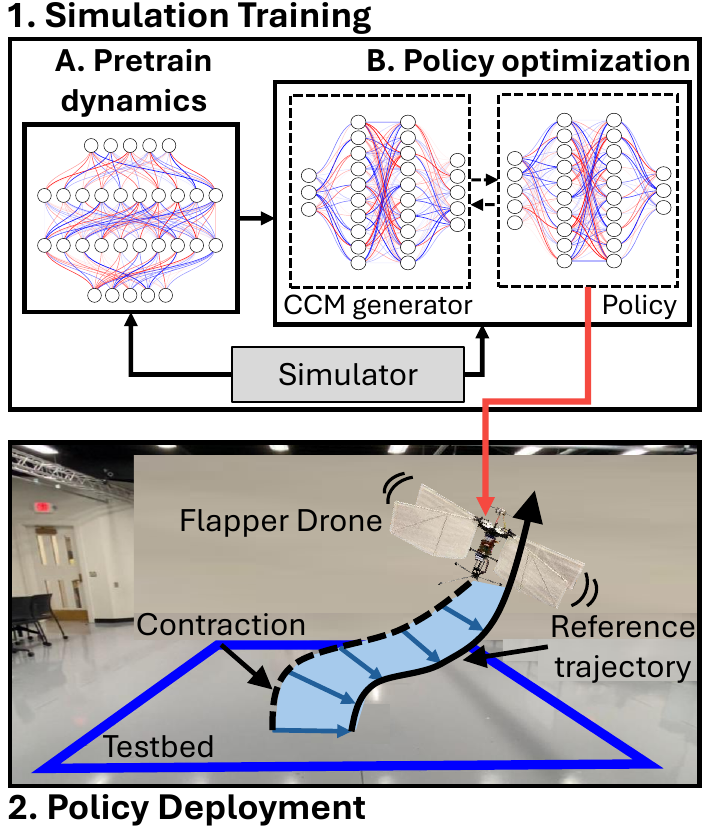}
    \caption{Our approach uses simulation-based training, where we first pretrain an approximate dynamics model for robots, which is then used to jointly learn a CCM generator and optimal policy. This policy is then deployed on the real robots for contraction-aware and robust path-tracking.}
    \label{fig:cac_intro}
\end{figure}

Sun et al. \cite{sun2021learning} adopt a learning-based approach to identify CCMs and synthesize contracting policies, thereby addressing the scalability and structural limitations of non-learning methods.
They jointly train two neural networks---one for a CCM generator (CMG) and another for a policy---by leveraging violations of CCM conditions as their training loss. 
However, this approach only ensures the pointwise satisfaction of CCM conditions and therefore does not consider long-term optimality (i.e., cumulative tracking error over an entire trajectory).
Furthermore, Richards et al. \cite{richards2023learning} show that such a myopic formulation could make the synthesized policy more susceptible to learning biases when approximate dynamics are used to model CCM conditions.

\cite{song2024stable} and \cite{zinage2026contractionppocertifiedreinforcementlearning} integrate reinforcement learning (RL) with contraction theory to provide a contraction certificate to RL-trained policies.
\cite{song2024stable} introduces a coordinate transformation to linearize nonlinear contraction constraints and apply these constraints to the neural network weights of the policy, and \cite{zinage2026contractionppocertifiedreinforcementlearning} introduces an auxiliary reward to the path-tracking reward to encourage satisfaction of the contraction condition.
However, the former does not account for long-term optimality or guarantee the existence of such a coordinate transformation, particularly in high-dimensional nonlinear systems, and the latter does not establish that the proposed reward function leads to the actual synthesis of a contracting policy.

In this paper, we integrate CCMs into RL in a manner that enables the synthesis of contraction-aware, robust policies that achieve long-term optimality for high-dimensional nonlinear dynamical systems with minimal structural assumptions.
Our approach is visualized in \Cref{fig:cac_intro}.
Our key idea is to use CCMs to define a reward function such that the optimal policy---trained using RL---exhibits both incremental asymptotic stability and long-term optimality.
We implement this idea through our new algorithm, contraction-aware RL (CARL), which jointly learns a mapping from the system state to CCMs (referred to as the CMG) and a policy that maximizes rewards defined by CCMs sampled from the CMG.
We use proximal policy optimization (PPO) \cite{schulman2017proximal} for RL optimization in this work, though in principle, our approach could be used with any RL algorithm.
We empirically validate that CARL improves path-tracking performance compared to established baselines in four simulated environments of increasing dimensionality, ranging from a 4D Car to a 10D Quadrotor system.
We also evaluate the robustness of CARL to approximation errors in the pretrained dynamics model (by considering simulated datasets of varying coverage) and to sim-to-real errors (by performing robot experiments with a TurtleBot3 Burger, a highly nonlinear Flapper Drone, and a high-dimensional complex Jethexa hexapod).
As for our theoretical rationale, we prove the incremental asymptotic stability of an optimal policy trained by our algorithm and formally characterize the trade-off between long-term optimality and incremental exponential stability.
We additionally provide conditions under which path-tracking policies, including those produced by CARL, are robust to approximation errors in the pretrained dynamics.

\subsection{Contributions}
Our contributions are summarized as follows:
\begin{enumerate}
    \item We propose a contraction-aware RL algorithm (CARL) that uses CCMs and RL to synthesize a robust path-tracking policy that enhances long-term optimality.

    \item We formally prove that the policy optimized by CARL exhibits incremental asymptotic stability and can also achieve incremental exponential stability by modulating the planning horizon governed by the discount factor, and establish conditions under which the optimized policy remains robust to approximation errors in the pretrained dynamics.
    
    \item We validate the advantages of prioritizing long-term optimality by CARL in both simulated environments and on diverse robotic platforms.
    
\end{enumerate}

\subsection{Organization}
We review related work in \Cref{sec:related_works} and provide preliminaries in \Cref{sec:preliminaries}. 
We present CARL in \Cref{sec:CAC} and provide a theoretical analysis in \Cref{sec:theory}. 
We present empirical results from simulated experiments in \Cref{sec:experiments} and three real-world robot experiments in \Cref{sec:experiment_ground_robot} (wheeled),  \Cref{sec:experiment_aerial_robot} (aerial), and \Cref{sec:experiment_legged_robot} (legged).
We discuss concluding thoughts, limitations, and future work in \Cref{sec:conclusion}.

\subsection{Notation}
Since this work combines concepts from contraction theory and RL, we provide a comprehensive summary of the notation used throughout the paper in \Cref{tab:notation}.
\begin{table}[t]
\centering
\caption{Summary of notation.}
\label{tab:notation}
\renewcommand{\arraystretch}{1.3}
\begin{tabular}{cl}
\toprule
\textbf{Symbol} & \textbf{Description} \\ 
\midrule
\multicolumn{2}{l}{\textit{Coordinates \& Kinematics}} \\
$p, v, a$ & Position, velocity, and acceleration \\
$\phi, \theta, \psi$ & Euler angles: roll, pitch, and yaw \\
$\tilde{m}, g, \tilde{f}$ & Mass, gravitational acceleration, and thrust \\
$\mc{I}$ & Moment of inertia \\
\midrule
\multicolumn{2}{l}{\textit{System Dynamics}} \\
$x \in \mathcal{X}, u \in \mathcal{U}$ & State and control with their associated spaces \\
$x^* \in \mathcal{X}, u^* \in \mathcal{U}$ & Reference state and control with their spaces \\
$\delta x$ & Differential state (virtual displacement) \\
$f(x), B(x)$ & Drift dynamics and actuation matrix \\
$\hat f_\xi (x), \hat B_\zeta(x)$ & Approximate $f$ and $B$ with parameters $\xi$ and $\zeta$ \\
\midrule
\multicolumn{2}{l}{\textit{Contraction Theory}} \\
$M(x)$ & Control contraction metrics \\
$\mc{V}(x, \delta x)$ & Differential Lyapunov function \\
\midrule
\multicolumn{2}{l}{\textit{Reinforcement Learning}} \\
$s \in \mathcal{S}$ & RL state $s_t = (x_t, x_{t:t+N-1}^*, u_t^*)$ \\
$N$ & A fixed look-ahead horizon \\
$P(s'|s,u)$ & Transition probability density function \\
$R(s,u)$ & Reward function \\ 
$\gamma$ & Discount factor \\
$\pi_\vartheta(u\mid s)$ & Stochastic policy with parameters $\vartheta$ \\
$G_t$ & Expected return, $G_t=\sum_{k=0}^\infty \gamma^{k} R(s_{t + k}, u_{t + k})$ \\ 
$V_\varphi^{\pi_\vartheta}(s)$ & State-value function for $\pi_\vartheta$ with parameters $\varphi$ \\ 
$Q^{\pi_\vartheta}(s,u)$ & Action-value function for $\pi_\vartheta$ \\
$d_0$ & Initial state distribution, $s_0 \sim d_0$ \\
$d^{\pi_\vartheta}(s,u)$ & Discounted state-control occupancy for policy $\pi_\vartheta$ \\

\bottomrule
\end{tabular}
\end{table}

\section{Related Works} \label{sec:related_works}

\subsection{Non-learning Synthesis of Contracting Policies}
Early works on CCMs focused on analytically identifying them in an offline phase and computing corresponding controls in an online phase.
For example, \cite{manchester2017control} uses the dual metric of CCMs with SoS programming to solve an infinite-dimensional convex feasibility problem with the CCM conditions.
Then, they define the differential feedback gain and path-integrate it to obtain the control that drives the system dynamics to contract.
\cite{singh2017robust} improves the method in \cite{manchester2017control} by additionally considering CCMs that minimize the area of the robust control invariant tube (i.e., they balance between contraction rate and overshoot) and using quadratic programming to find the most efficient control (i.e., control with lowest $\ell_2$-norm) for system contraction.
However, these approaches require extensive computation to minimize geodesics, which can be computationally intractable for real-time deployment.
Furthermore, the computational complexity of the SoS optimization scales exponentially with the problem dimension, restricting its use to small-scale, low-dimensional applications.

To address the intractability of computing exact geodesics while robustly accounting for external disturbances, \cite{tsukamoto2020robust} proposes the ConVex optimization-based Stochastic steady-state Tracking Error Minimization (CV-STEM) framework. 
This framework identifies contraction metrics that minimize an upper bound on the steady-state mean-squared tracking error of the system, and then uses them to derive a nonlinear counterpart of the linear quadratic regulator (LQR) feedback law. 
This controller design guarantees exponential boundedness of the tracking error and exhibits $L_2$-robustness properties, while avoiding the need to compute minimizing geodesics.
While this approach is more tractable than prior work, it remains computationally challenging for hardware-constrained robots. 
To further mitigate this real-time computational burden, \cite{tsukamoto2021learning} proposes a two-stage approach: they first compute the contraction metric and controls using CV-STEM in an offline phase, and subsequently employ supervised learning to distill these controls into a neural network policy for efficient real-time deployment.
However, this approach requires a dense dataset for successful distillation and may not scale to high-dimensional systems, as finding contraction metrics is time-consuming and the volume of data required for successful distillation increases with state dimension.

While contraction theory offers a rigorous framework for nonlinear stability, classical linear control approaches remain standard for practical path-tracking applications. 
The most prominent example is LQR \cite{kalman1960contributions}, which addresses the path-tracking problem by finding an optimal constant gain matrix with respect to the linearized error dynamics. 
Given that LQR relies on linearization of error dynamics and known system dynamics, Richards et al. \cite{richards2023learning} propose to approximate the system dynamics using neural networks and apply the state-dependent coefficient factorization (i.e., $f(x) = A(x) x$) to formulate the exact nonlinear error dynamics.
These error dynamics are then used to solve the state-dependent Riccati equation in real-time to find stabilizing controllers. 
This approach has been shown to achieve a larger region of attraction (RoA) than standard LQR \cite{cimen2012survey, richards2023learning}. 
However, \cite{richards2023learning} lacks formal guarantees for certifying the exact size of the RoA and is only locally stabilizing for nonlinear system dynamics.
Additionally, LQR-based approaches require computationally intensive real-time optimization (as we show in \Cref{table:mauc_report}).

See \cite{tsukamoto2021contraction} and the references therein for a comprehensive survey of these, and other, non-learning methods for synthesizing contracting policies.

\subsection{Learning-based Synthesis of Contracting Policies}
To mitigate the efficiency, scalability, and stability limitations of non-learning approaches, Sun et al. \cite{sun2021learning} propose a gradient-based learning approach for synthesizing policies with CCMs.
More specifically, their approach learns neural networks that model a CCM generator and corresponding contracting policy.
To train the neural networks, they first use a known dynamics model to evaluate the violation of CCM conditions for current outputs from the CMG and policy.
Then, these violations are used as a loss function for simultaneously updating the CMG and policy. 
While this approach addresses efficiency, scalability, and stability limitations, it only focuses on satisfying the CCM conditions in a pointwise manner, which again does not ensure long-term optimality.

\cite{song2024stable} recently proposed a modular control architecture that guarantees exponential stability within RL frameworks.
They use a coordinate transformation to linearize nonlinear contraction constraints into algebraically solvable linear constraints, and apply these constraints to the weights of the policy's neural network.
Specifically, they restrict the signs of the neural network's weights (e.g., to be strictly positive or negative) to satisfy algebraic constraints, ensuring the policy generates controls that drive the system to contract.
However, the existence of the necessary coordinate transformation is not guaranteed, which may limit the method's applicability to general high-dimensional nonlinear systems.
\cite{zinage2026contractionppocertifiedreinforcementlearning} employs PPO to synthesize a contracting policy by incorporating a contraction penalty as an auxiliary reward. 
They also analyze the worst-case contraction rate under the dynamics mismatch from a sim-to-real transfer. 
However, they do not provide a formal justification for how optimizing this augmented reward function leads to the actual synthesis of a contracting policy.

\section{Preliminaries} \label{sec:preliminaries}
\subsection{Dynamical System}\label{subsec:dynamical_system}
We consider control-affine nonlinear systems given by
\begin{equation} \label{eqn:control_affine_system}
    \dot{x}(t) = f(x(t)) + B(x(t)) u(t),
\end{equation}
where $x(t) \in \mathcal{X} \subset \mathbb{R}^n$ with $\mathcal{X}$ a compact state set, $u(t)\in \mc{U} \subset \mb{R}^m$ with $\mc{U}$ a compact control set, $ f : \mb{R}^n \rightarrow \mb{R}^n $ denotes the smooth nonlinear drift dynamics, and $ B : \mb{R}^n \rightarrow \mb{R}^{n \times m} $ is the smooth actuation matrix function.
For notational simplicity, we omit the explicit time dependence to write $ x(t) $ and $ u(t) $ as $ x $ and $ u $, unless the time argument is required for clarity.

\subsection{Contraction Theory and Control Contraction Metrics}
Contraction theory \cite{LOHMILLER1998683} provides a framework for analyzing the convergence between solution trajectories satisfying \Cref{eqn:control_affine_system} by leveraging the differential form of Lyapunov functions. Specifically, contraction theory reformulates Lyapunov stability conditions by employing a quadratic Lyapunov function defined over differential dynamics as follows:
\begin{equation} \label{eqn:differential_lyapunov_function}
    \mc{V}(x(t), \delta x(t)) = \delta x(t)^\top M(x(t)) \delta x(t),
\end{equation}
where $ \delta x(t) $ is the differential state (i.e., virtual displacement) between two solution trajectories at time $t$, and $ M(x(t)) \succ 0 $ is the symmetric positive definite matrix defining the contraction metric. When the time derivative of $\mc{V}$ satisfies $\dot{\mc{V}} \leq -2\lambda \mc V$ for some $\lambda > 0$, the differential dynamics of $\delta x(t)$ are guaranteed to decay exponentially at rate $\lambda$, which implies incremental exponential stability of the system.
Specifically, let the contraction metric be bounded by $\underline{m}I \preceq M \preceq \overline{m}I$, where $\overline{m}, \underline{m} \in \mathbb{R}$ satisfy $0<\underline{m}\le\overline{m}$ and $I$ is the identity matrix.
Then, the following exponential stability holds:
\begin{equation}\label{eqn:euclidean_contraction}
    \| \delta x(t) \|_2 \leq \sqrt{\frac{\overline{m}}{\underline{m}}} \|\delta x(t_0)\|_2 \cdot e^{-\lambda (t - t_0)},
\end{equation}
where $\sqrt{{\overline{m}} / {\underline{m}}}$ accounts the overshoot.

\cite{manchester2017control} extends contraction theory to constructive closed-loop nonlinear control design. Formally, the system in \Cref{eqn:control_affine_system} with a control policy $u(t) = \pi_c(x(t),t)+v(t)$ (where $v(t)$ is an arbitrary function) is said to be contracting if there exists a matrix $ M(x) \succ 0 $ that satisfies
\begin{equation}\label{eqn:contraction_condition}
    \dot{M} + 2 \text{sym}\left( M(A + BK) \right) + 2 \lambda M \preceq 0,
\end{equation}
where we omit the input argument for clarity, sym$(\cdot)$ is the symmetric part of the input argument (i.e., $\text{sym}(A)=(A + A^\top)/2$), $ A(x, u) = \frac{\partial f}{\partial x} + \sum_{i=1}^m u_i \frac{\partial b_i}{\partial x} $ is the Jacobian of the drift dynamics, $ K(x, t) = \frac{\partial \pi_c}{\partial x} $ represents the differential feedback gain associated with the policy, and $ \lambda > 0 $ specifies the contraction rate. A sufficient condition for the existence of a contracting policy satisfying \Cref{eqn:contraction_condition} is given as follows:
\begin{align}
    B_\perp^\top \big( 
        -\partial_f W(x) 
        + 2\text{sym}\big( \tfrac{\partial W(x)}{\partial x} f(x) \big) 
        + 2\lambda W(x) 
    \big) B_\perp \prec 0,
    \label{eqn:Bperp_condition1} \\
    B_\perp^\top \big( 
        \partial_{b_j} W(x) 
        - 2\text{sym}\big( \tfrac{\partial b_j(x)}{\partial x} W(x) \big) 
    \big) B_\perp = 0,
    \; j = 1, \dots, m.
    \label{eqn:Bperp_condition2}
\end{align}
Here, $W(x)= M^{-1}(x)$ denotes the dual metric of $M$, $\partial_f W(x)=\sum_{i=1}^n f_i \frac{\partial W}{\partial x_i}$ represents the Lie derivative of the matrix-valued function $W(x)$ along the vector $f \in \mb{R}^n$, and $B_\perp$ denotes the matrix whose columns span the null space of the actuation matrix $B^\top$. 
The conditions above ensure that the dynamics evolving orthogonally to the span of actuated directions (i.e., directions that cannot be controlled by $u$) must naturally contract.
A matrix $M$ that satisfies \Cref{eqn:contraction_condition,eqn:Bperp_condition1,eqn:Bperp_condition2} defines a control contraction metric, and the satisfaction of these conditions ensures that the policy is guaranteed to exhibit incremental exponential convergence to the reference trajectory, serving as a strong certificate of stability.

\subsection{Reinforcement Learning}
Reinforcement learning (RL) addresses sequential decision-making under uncertainty through interactions with environments. In this work, we formulate the path-tracking problem as a Markov decision process (MDP) \cite{puterman2014markov}.

\subsubsection{Markov decision process} 

An MDP is defined by a tuple $ ( \mathcal{S}, \mathcal{U}, P, R, \gamma ) $, where $\mathcal{S}$ is the state space, $\mathcal{U}$ is the control space, $ P: \mathcal{S} \times \mathcal{U} \times \mathcal{S} \rightarrow [0,\infty) $ is the probability density of the next state being $s' \in \mathcal{S}$ given the current state $s \in \mathcal{S}$ and control input $u \in \mathcal{U}$, $ R: \mathcal{S} \times \mathcal{U} \rightarrow \mb{R} $ is the reward function, and $ \gamma \in [0,1) $ is the discount factor.
At time step $t$, the agent executes control $u_t \in \mathcal{U}$ given the current state $s_t \in \mathcal{S}$, after which the system transitions to state $s_{t+1} \in \mathcal{S}$ and the agent receives reward $r_t = R(s_t, u_t)$. 
The expected return at time $t$ is defined as $ G_t = \sum_{k=0}^{\infty} \gamma^k r_{t+k} $.

For our path-tracking problem, we define the state space as $\mc{S}=\mc{X} \times\mc{X}^N\times\mc{U}$, where $\mathcal{X}$ and $\mathcal{U}$ are defined as in \Cref{subsec:dynamical_system}, and $N$ is a fixed look-ahead horizon.
The state at time $t$ is defined by the agent state, the reference state, and the reference control such that $s_t=(x_t, x_{t:t+N-1}^\ast, u_t^\ast)$.
Our goal is to optimize a policy $\pi: \mc{S} \times \mc{U} \rightarrow [0, \infty)$, which defines a probability density over continuous controls $\mathcal{U}$ and maximizes the expected return.

\subsubsection{Policy Optimization}
Policy gradient methods are commonly used for policy optimization when considering high-dimensional and continuous state and control spaces.
Let $\pi_\vartheta$ be a policy parameterized by $\vartheta$ and $V^{\pi_{\vartheta}}(s) = \mathbb{E}_{\pi_{\vartheta}} [G_t \mid s_t = s]$ be its corresponding state-value function.
The goal is to maximize the following performance objective, where $d_0: \mathcal{S} \rightarrow [0, \infty)$ is the initial state distribution:
\begin{equation}\label{eqn:performance_of_policy}
    J(\vartheta) = \mathbb{E}_{s_0 \sim d_0} \left[ V^{\pi_\vartheta}(s_0) \right].
\end{equation}

Proximal policy optimization (PPO) \cite{schulman2017proximal} is a widely used policy optimization algorithm based on an actor-critic architecture that has shown strong performance and efficiency across a wide range of robotic tasks \cite{tang2025deep}. 
In this work, we employ PPO for policy optimization, which often includes entropy, denoted by $\mc{H}(\pi_\vartheta(\cdot \mid s))$, in the performance objective as follows:
\begin{equation}\label{eqn:PPO_entropy_regularization}
    J(\vartheta) \leftarrow J(\vartheta) + \beta_\pi \mb{E}_{s\sim d^{\pi_{\vartheta}}} \left[ \mc{H}(\pi_{\vartheta} (\cdot \mid s)) \right],
\end{equation}
where $\beta_\pi > 0$ is a hyperparameter and $d^\pi$ is the discounted state occupancy induced by $\pi$.
The entropy regularization is known to smooth the optimization landscape \cite{ahmed2019understanding} by encouraging exploration of a policy.

\section{Main Result: An Algorithm for Contraction-Aware Reinforcement Learning} \label{sec:CAC}
\begin{figure*}[t]
    \centering
    \includegraphics[width=0.99\linewidth]{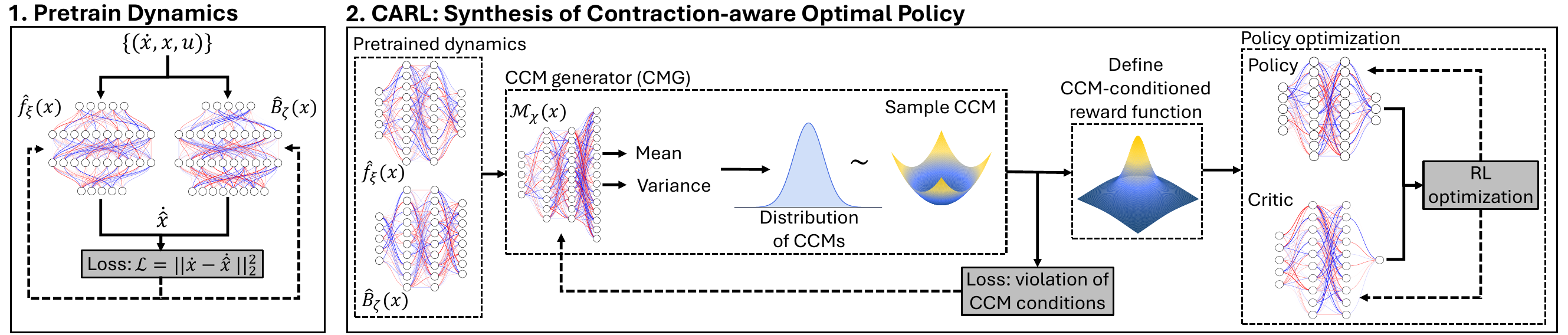}
    \caption{
    We present a comprehensive overview of our proposed algorithm. We first pretrain a dynamics model using an offline dataset that contains tuples of $(\dot x, x, u)$. Then, we jointly train a CCM generator (CMG), using the pretrained dynamics model, and an RL policy, using a CCM-conditioned reward function that is evaluated by sampling the CMG. This formulation allows the RL policy to minimize cumulative tracking error with a stability guarantee.
    }
    \label{fig:cac_algorithm}
\end{figure*}

We propose a new algorithm, contraction-aware RL (CARL), for synthesizing a robust path-tracking policy that achieves incremental asymptotic stability and long-term optimality. 
CARL is visualized in \Cref{fig:cac_algorithm} and outlined in \Cref{alg:cac}.
It is composed of two primary steps: pretraining a dynamics model and then jointly learning a CCM generator (CMG) and policy.

Our empirical results (see \Cref{sec:experiment_ground_robot,sec:experiment_aerial_robot,sec:experiment_legged_robot}) demonstrate that the resulting policy can then be deployed on real-world robots in a zero-shot manner. We describe each step of CARL in more detail below.

\begin{algorithm}[t]
    \caption{Contraction-Aware RL ($\text{CARL}$)}
    \label{alg:cac}
    \begin{algorithmic}[1]
        \REQUIRE Offline data $\mc{D} = \{ (\dot x_i, x_i, u_i) \}_{i=1}^{N_{\text{data}}}$, state $\mc{X}$, control space $\mc{U}$, $N_\text{dynamics}$, $N_\text{CMG}$, $N_\text{policy}$
        \item[\textbf{Initialize:}] Neural network models $\hat{f}_\xi, \hat{B}_\zeta, \mc{M}_\chi$, $\pi_\vartheta$, and $V^{\pi_\vartheta}_\varphi$

        \STATE \texttt{/* Pretrain dynamics model */}
        \FOR{iteration = 1, 2, \dots, $N_{\text{dynamics}}$}
            \STATE Sample batch $d \sim \text{Unif}(\mc{D})$
            \STATE Update dynamics parameters $ \xi, \zeta$ with Eqn. \ref{eqn:dynamic_learning_loss} using $d$
        \ENDFOR
        
        \STATE \texttt{/* Policy optimization */}
        \FOR{iteration = 1, 2, \dots, $N_{\text{CMG}}$}
            \STATE Sample batch $d \sim \text{Unif}(\mc{X},\mc{U})$
            \STATE Update CMG parameter $ \chi $ with Eqn. \ref{eqn:cmg_loss} using $d$
            \FOR{iteration = 1, 2, \dots, $N_{\text{policy}}$}
                \STATE Collect rollouts using current policy $\pi_\vartheta$.
                \STATE Calculate CCM-conditioned rewards using CCMs $M = \mc{M}_\chi(x)$
                \STATE Update policy parameters $ \vartheta $ and critic parameters $\varphi$
            \ENDFOR
        \ENDFOR     
    \end{algorithmic}
\end{algorithm}

\subsection{Pretraining a Dynamics Model}
\label{sec:pretraining-dynamics-model}
We first pretrain an approximate dynamics model, which we later use to train a CMG (see \Cref{sec:jointly-learning-cmg-policy}).
We approximate the dynamics by learning approximations of the drift dynamics $f$, the actuation matrix function $B$, and the annihilator matrix of the actuation matrix $B_\perp$. Following Richards et al. \cite{richards2023learning}, we employ neural networks to model $f$ and $B$ and use singular value decomposition (SVD) to compute $ B_\perp $. Specifically, given a dataset $ \mathcal{D} = \{ (\dot{x}_i, x_i, u_i) \}_{i=1}^{N_{\text{data}}} $, we simultaneously learn two feed-forward networks, $ \hat{f}_{\xi} $ and $\hat{B}_{\zeta}$, parameterized by $\xi$ and $\zeta$ respectively, as follows:
\begin{equation}\label{eqn:dynamic_learning_loss}
    \xi^\star, \zeta^\star = \argmin_{\xi, \zeta} \, \mb{E}_{(\dot{x}, x, u) \sim \mathcal{D}} \left[ \left\| \dot{x} - \left( \hat{f}_\xi(x) + \hat{B}_\zeta(x) u \right) \right\|^2_2 \right].
\end{equation}
Then, the annihilator matrix of $\hat{B}^\top$ (i.e., $\hat{B}_\perp$) is computed using SVD, as it is spanned by the last $n - \mathrm{rank}(\hat{B})$ columns of $\mb{U}$ in $\text{SVD}(\hat{B}) = \mb{U} \Sigma \mb{V}^\top$. For instance, for $\hat{B} \in \mathbb{R}^{3 \times 2}$ with $\mathrm{rank}(\hat{B}) = 2$, the third column of $\mb{U} \in \mathbb{R}^{3 \times 3}$ spans the null space of $\hat{B}^\top$, which we use as the approximation of the annihilator matrix $\hat{B}_\perp$.
We employ SVD to leverage existing code libraries, though a naive Gram–Schmidt process is a valid alternative for constructing the annihilator matrix.


\subsection{Jointly Learning a CMG and a Policy}
\label{sec:jointly-learning-cmg-policy}
Given a pretrained dynamics model, we then simultaneously learn a CMG and a policy that maximizes rewards defined by the CMG.
This process can be numerically unstable because the CMG is conditioned on the current policy.
We address this issue by implementing a freeze-and-learn strategy, where for each update of the CMG, we freeze its parameters and then perform several policy updates to allow the RL policy to adapt to the newly defined reward landscape.
Note that the pretrained dynamics model is only used to train the CMG; it is not directly used for policy optimization.

\subsubsection{Learning a control contraction metric generator} 
\label{sec:cmg}
Given a pretrained dynamics model, we train a CMG $\mathcal{M}_\chi$, parameterized by $\chi$, to generate CCMs for the current policy. 
Recall that CCMs are $n \times n$ matrices satisfying \Cref{eqn:contraction_condition,eqn:Bperp_condition1,eqn:Bperp_condition2}, where $n$ is the dimension of the state $x$, as defined in \Cref{subsec:dynamical_system}.
We assume a probabilistic model for the CMG, where CCMs are modeled as a random matrix with each element being an independent univariate Gaussian distribution.
The CMG is then trained to output the mean and variance for each of those distributions---that is, the CMG outputs $n^2$ mean values and $n^2$ variances.
We then sample a CCM for the current policy and state $x$ as $M \sim \mathcal{M}_\chi(\cdot \mid x)$.
Inspired by \cite{ahmed2019understanding, sun2021learning}, we use this probabilistic model for the CMG to train it with entropy regularization, which we use to encourage exploration in finding feasible CCMs and prevent premature convergence.

We train the CMG using the loss function from Sun et al. \cite{sun2021learning} with an additional value-conditioned entropy regularizer.
Specifically, let $ C_M(\cdot) $, $ C_{W_1}(\cdot) $, and $ C_{W_2}^j(\cdot) $ denote the left-hand side of \Cref{eqn:contraction_condition,eqn:Bperp_condition1} and the $j$-th equality constraint in \Cref{eqn:Bperp_condition2} for $ j = 1, \dots, m $, respectively. Then, the CMG is trained to minimize the violation of CCM conditions over the state and control spaces as follows:
\begin{equation}\label{eqn:cmg_loss}
    \begin{aligned}
        \chi^\star = \argmin_\chi \; &  \mathbb{E}_{\substack{x\sim \mathrm{Unif}(\mathcal{X}),\\ u \sim \pi_\vartheta(\cdot \mid x),\\ M\sim\mathcal{M}_\chi(\cdot \mid x),\\
        (x^*, u^*) \sim \mc{T}}}
  \Bigl[ \mc{L}(M - \overline{m}I) + \mc{L}(C_M(M, x,u)) \\
  & + \mc{L}(C_{W_1}(M^{-1},x)) + \sum_{j=1}^m \|C^j_{W_2}(M^{-1},x)\|_F  \\
  & - \alpha(V^{\pi_\vartheta}(s)) \cdot \mc{H}(\mc{M}_\chi(\cdot \mid x))\Bigr].
    \end{aligned}
\end{equation}
Here, $\mc{T}(x^*,u^*)$ is a distribution over a set of reference trajectories and $\mc{L}(A) = (z^\top A z) /(z^\top z) $ if $A \succ 0$ else $0$ is a loss function that penalizes positiveness of the input matrix, where $z\in\mb{R}^n$ and $z_i \sim \text{Unif}(-1,1)$. Additionally, $||\cdot||_F$ is the Frobenius matrix norm and $\mc{H}(\mc{M}_\chi(\cdot \mid x))$ is the entropy of the probability distribution $\mc{M}_\chi$.
The strength of the entropy regularizer is modulated by the value function $V^{\pi_\vartheta}(s)$,
where $\alpha(V^{\pi_\vartheta}(s)) = \beta_{\mc{M}} e^{-V^{\pi_\vartheta}(s)}$ with hyperparameter $\beta_\mc{M} >0$. Intuitively, the value-conditioned entropy regularizer reduces the degree of exploration by the CMG when the agent achieves high values and increases it when the agent achieves low values.

The auxiliary loss terms $\mathcal{L}(C_{W_1}(\cdot))$ and $\| C^j_{W_2}(\cdot) \|_F$ in \Cref{eqn:cmg_loss} are critical for convergence of the CMG, as directly optimizing $\mathcal{L}(C_M(\cdot))$ has been shown to struggle with convergence due to the bilinear coupling between CCMs and the policy~\cite{sun2021learning}. 
Specifically, the CMG requires a stabilizing policy to determine accurate update directions, while the policy conversely relies on a high-fidelity CMG to guide its own optimization. 
The auxiliary loss terms resolve this bilinear coupling by guiding the CMG toward a structurally valid metric space that is independent of the policy.
However, strict minimization of these auxiliary loss terms is not required because they represent sufficient, but not necessary, conditions for the contraction of the dynamics.
Consequently, these terms function as regularizers for stable training.

In \Cref{eqn:cmg_loss}, we additionally reduce the overshoot that is proportional to $\sqrt{\overline{m}/\underline{m}}$ (see \Cref{eqn:euclidean_contraction}).
Specifically, we ensure the eigenvalues of $M$ are within the range of $\underline{m} $ and $ \overline{m}$ by adding $\underline{m}$ to the generated CCMs (i.e., $M = \mc{M}_\chi(x) + \underline{m}I$) and penalizing the violation of the upper bound through $ \mc{L}(M - \overline{m}I)$.

\subsubsection{Learning a control contraction metric-guided policy} 

Prior works that synthesize contracting policies (see \Cref{sec:related_works}) do not consider long-term optimality.
Specifically, suppose there exist CCMs that satisfy the conditions in \Cref{eqn:contraction_condition,eqn:Bperp_condition1,eqn:Bperp_condition2}. This guarantees that there exists a set of control inputs $\tilde{\mathcal{U}} \subseteq \mc{U}$ that ensures contraction \cite{manchester2017control}. While multiple choices of control $u\in\tilde{\mc{U}}$ may satisfy the CCM conditions under a given metric, they are not guaranteed to minimize the cumulative tracking error across the entire system trajectory.
The key idea of CARL is to instead use the CMG to define a CCM-conditioned reward function, which we then optimize using RL to synthesize path-tracking policies that ensure incremental asymptotic stability and long-term optimality.

We specifically use the following CCM-conditioned reward function:
\begin{equation}\label{eqn:our_reward_function}
    R(s) \coloneq \frac{1}{1 + \delta x^\top M\delta x} \in [0, 1],
\end{equation}
where we approximate the infinitesimal displacement, $\delta x$, as the Euclidean displacement between $x$ and $x^*$ (i.e., $\delta x = x  - x^*$) as in \cite{sun2021learning}, since computing exact geodesics is computationally prohibitive. This approximation is valid because the reward approaches its maximum value of $1$ upon contraction (i.e., $\delta x = 0$) and approaches $0$ as $\delta x \rightarrow \infty$, effectively encouraging contraction.
Additionally, we include the $+1$ term in the denominator to ensure the reward is bounded within $[0,1]$, motivated by prior work that has shown bounded rewards can improve numerical stability of RL algorithms \cite{engstrom2019implementation, andrychowicz2021matters, shengyi2022the37implementation}.
We deterministically generate $M$ by defining it as the mean from the CMG $\mc{M}_\chi(x)$, avoiding the complexity of training with a stochastic reward function since a single valid CCM for each state is sufficient for contraction.

Our proposed reward function achieves two key properties.
First, the reward uses CCMs to ensure the resulting optimization landscape is contraction-aware---that is, the reward function in \Cref{eqn:our_reward_function} encourages convergence to policies that not only minimize cumulative tracking, but also do so in a way that contracts.
Second, the reward is defined such that its maximization is equivalent to minimizing the cumulative tracking error, thus ensuring long-term optimality.
We prove this equivalence in \Cref{sec:stability-guarantee}.

While the reward defined in \Cref{eqn:our_reward_function} is intuitively and theoretically justified (see \Cref{sec:theory} for theory), in practice, we include an entropy regularization term, again motivated by prior work showing the benefit of entropy regularization for exploration in RL \cite{ahmed2019understanding}.
We also deterministically execute the trained stochastic policies for evaluation using the mean of the control distribution to better align our theoretical analysis (\Cref{sec:theory}), which assumes deterministic policies.
\cite{jang2025stochastic} also empirically shows that deterministic policy evaluation generally outperforms stochastic evaluation.

\section{Theoretical Results: \\Bridging RL and Contraction Theory} \label{sec:theory}

This section provides theoretical justification for CARL. 
In \Cref{sec:stability-guarantee}, we first prove the incremental asymptotic stability of an optimal policy for our CCM-conditioned reward, and demonstrate how the discount factor modulates the trade-off between long-term optimality and incremental exponential stability.
In \Cref{sec:robustness_theory}, we then establish conditions under which that policy remains robust to approximation errors in the pretrained dynamics.

We consider an optimal policy $\pi^\star$ that maximizes the expected return for the reward defined in \Cref{eqn:our_reward_function} (i.e., without entropy regularization) to study the core idea of a CCM-conditioned reward.
We also restrict our analysis to deterministic optimal policies to align with the deterministic formulation of CCMs \cite{manchester2017control}.
This restriction is justified without loss of generality, since a deterministic stationary optimal policy is guaranteed to exist for infinite-horizon discounted MDPs with compact state-control spaces and smooth dynamics \cite{puterman2014markov, bertsekas1996stochastic}.
Additionally, \cite{montenegro2024learning} theoretically shows that optimizing stochastic policies with proper annealing of stochasticity can converge to an optimal deterministic policy.
All proofs are deferred to Appendix \ref{app:proofs}.

We assume the following throughout our analysis.

\begin{assumption}\label{assumption:1}
    There exists at least one deterministic contracting policy $\pi_c$ that satisfies \Cref{eqn:contraction_condition} with a rate of $\lambda>0$. Hence, this policy $\pi_c$ satisfies
    \begin{equation}\label{eqn:contraction_inequality}
        \|\delta x(t_k)\|_M^2 \leq \|\delta x(t_0)\|_M^2 e^{-2\lambda k \Delta t}, \; \forall k \in \mb{N},
    \end{equation}
    where $\Delta t$ is a discrete time interval, $t_0$ is the initial time, $t_k = t_0 + k\Delta t$ is the current time, and $ \| \delta x(t) \|_M^2 = \int_0^1 \frac{\partial \tilde{x}}{\partial \mu}^\top M(\tilde{x}(\mu, t)) \frac{\partial \tilde{x}}{\partial \mu} \, d\mu $ denotes the integration over the geodesics connecting $ x(t) $ and $ x^*(t) $, with $ \tilde{x}(0, t) = x(t) $ and $ \tilde{x}(1, t) = x^*(t)$.
\end{assumption}

\subsection{Stability Analysis of a CCM-Conditioned Optimal Policy}
\label{sec:stability-guarantee}
Let us define a cost function as the path-tracking error under CCMs and its corresponding cumulative path-tracking error under a contracting policy $\pi_c$ as follows:
\begin{equation}\label{eqn:performance_measure_of_contracting_policy}
    C(s_t) \coloneq \| \delta x(t_0) \|_M^2, \quad V_C^{\pi_c}(s_t) \coloneqq \sum_{k = 0}^{\infty} \gamma^k C(s_{t+k}),
\end{equation}
and define the value function and objective for the reward function $R$ in \Cref{eqn:our_reward_function} for a policy $\pi$ as follows:
\begin{align}
    V_{R}^{\pi}(s) &  \coloneq \mb{E}_\pi \left[ \sum_{k=0}^{\infty} \gamma^k R(s_{t + k}) \mid s_t = s \right], \label{eqn:reward_value_function}\\
    J_{R}(\pi) & \coloneq \mb{E}_{s_0\sim d_0}\left[ V_{R}^{\pi}(s_0) \right] \label{eqn:reward_performance}.
\end{align}
Let $\pi^\star$ be a deterministic optimal policy for $J_{R}$ as follows:
\begin{equation}\label{eqn:discounted_optimal_policy}
    \pi^\star \coloneq \argmax_\pi J_{R}(\pi).
\end{equation}

To facilitate an analysis within a minimization framework, we introduce another cost function $\tilde{C}$ in terms of $R$ and its corresponding value function and objective as follows:
\begin{align} 
    \tilde{C}(s) & \coloneq 1 - R(s) \in [0, 1], \label{eqn:new_cost_function} \\
    V^\pi_{\tilde{C}}(s) & \coloneq \mb{E}_{\pi}\left[ \sum^\infty_{k=0} \gamma^k \tilde{C}(s_{t+k})  \mid s_t = s \right] \label{eqn:new_cost_value_function}, \\
    J_{\tilde C}(\pi) & \coloneq \mb{E}_{s_0\sim d_0}\left[ V^\pi_{\tilde{C}}(s_0) \right] \label{eqn:new_cost_performance}.
\end{align}
Then, we show the equivalence of minimizing the cost objective $J_{\tilde{C}}(\pi)$ and maximizing the reward objective $J_{R}(\pi)$.
\begin{lemma}\label{lemma:equivalence_of_optimizations}
\emph{(Equivalence of Cost Minimization and Reward Maximization).}
Let the reward and cost objectives be defined as in \Cref{eqn:reward_performance} and \Cref{eqn:new_cost_performance}, respectively.
Then, for a discount factor $\gamma \in [0, 1)$, the following equivalence holds
\begin{equation}\label{eqn:cost_minimization_objective}
    \begin{aligned}
        \pi^\star = \argmin_\pi J_{\tilde{C}}(\pi) = \argmax_\pi J_{R}(\pi),
    \end{aligned}
\end{equation}
assuming consistent tie-breaking rules.
\end{lemma}

We also establish the contraction of $\pi_c$ with respect to $\tilde{C}$.
\begin{lemma}\label{lemma:contraction_in_C_tilde}
    \emph{(Contraction with respect to $\tilde{C}$).}
    Suppose there exists at least one policy $\pi_c$ that contracts with respect to $C$ with rate $\lambda >0$ as in \Cref{assumption:1}.
    Then, there exists a rate $\tilde{\lambda} \in (0, \lambda)$ such that $\pi_c$ contracts with respect to $\tilde{C}$ as follows:
    \begin{equation}
        \tilde{C}(s_{t+k}) \leq \tilde{C}(s_t) e^{-2\tilde{\lambda} k \Delta t}, \; \forall k \in \mb{N}.
    \end{equation}
\end{lemma}

The reward value function $V^\pi_R$, reward objective $J_R(\pi)$, and \Cref{lemma:equivalence_of_optimizations} require a discount factor $\gamma < 1$ for finiteness under an infinite horizon, which poses a challenge for stability analysis of typical RL algorithms.
However, due to our integration of CCMs with RL, we are able to establish that the optimal policy $\pi^\star$ satisfying \Cref{eqn:discounted_optimal_policy} guarantees incremental asymptotic stability for any discount factor $\gamma \in (0, 1)$.

\begin{theorem}\label{thm:asymptotic_stability}
    \emph{(Incremental Asymptotic Stability).}
    Suppose there exists at least one policy $\pi_c$ that contracts with respect to $C$ with rate $\lambda >0$ as in \Cref{assumption:1}.
    Then, for any discount factor $\gamma \in (0, 1)$, a deterministic optimal policy $\pi^\star$ satisfying \Cref{eqn:discounted_optimal_policy} renders the closed-loop system incrementally asymptotically stable:
    \begin{equation}
        \lim_{k\to\infty} \| \delta x(t_k)\|_M^2 = 0, \quad \forall \delta x(t_0)\in\mb{R}^n.
    \end{equation}
\end{theorem}

Finally, we establish incremental exponential stability as follows.
\begin{theorem}\label{thm:exponential_stability}
    \emph{(Incremental Exponential Stability).}
    Suppose there exists at least one policy $\pi_c$ that contracts with respect to $C$ with rate $\lambda >0$ as in \Cref{assumption:1}.
    Then, for a discount factor $\gamma < \min(\overline{\gamma}, 1)$, there exists a contraction factor $\overline{\rho} \in (0, 1)$ such that a deterministic optimal policy $\pi^\star$ satisfying \Cref{eqn:discounted_optimal_policy} renders the closed-loop system incrementally exponentially stable:
    \begin{equation}
        \| \delta x(t_k) \|_M^2 \leq \| \delta x(t_0) \|_M^2 \overline{\rho}^{k}, \; \forall k \in \mb{N},\;  \delta x(t_0)\in\mb{R}^n,
    \end{equation}
    where $\overline{\gamma} \coloneq e^{2\tilde{\lambda}\Delta t} - 1 > 0$ and $\tilde{\lambda}$ is the contraction rate induced by $\pi_c$ with respect to the cost function $\tilde{C}$ as in \Cref{lemma:contraction_in_C_tilde}.

\end{theorem}

\Cref{thm:exponential_stability} demonstrates that CARL can directly trade-off between long-term optimality and the strength of its stability guarantee through the modulation of the discount factor $\gamma$.
This is intuitive, since minimizing future errors (i.e., a long-term optimal strategy) may sacrifice near-term performance, causing overshoots that violate exponential stability.
If the threshold $\overline{\gamma} \geq 1$, \Cref{thm:exponential_stability} guarantees that the optimal policy $\pi^\star$ is both long-term optimal and exponentially stabilizing for discount factor $\gamma \rightarrow 1$.
Note that \Cref{thm:asymptotic_stability} and \Cref{thm:exponential_stability} are direct results of using our CCM-integrated reward function.

\subsection{Robustness of RL-trained Optimal Policy}
\label{sec:robustness_theory}
Finally, we analyze the robustness of $\pi^\star$ under the approximate dynamics model learned from \Cref{eqn:dynamic_learning_loss}. To this end, let us define the dynamics modeling error as
\begin{align}
    \mf{d}(x,u) \coloneq f(x) + B(x)u - \hat{f}_\xi(x) - \hat{B}_\zeta(x) u.
\end{align}
Also, for any trajectory segment of length $\kappa$, we define the truncated $\ell^2$ norm weighted by the contraction metric as
\begin{align}
    \|(\delta x)_\kappa\|_{\ell^2_M} \coloneq 
    \sqrt{\sum_{k=0}^{\kappa-1} \|\delta x(t_k)\|^2_M}.
\end{align}
The following theorem provides an extension of the result presented in~\cite{sihang_conformal}.

\begin{theorem}\emph{(Robustness of RL-trained Policies).}
    \label{thm_robustness}
    Suppose that both the contracting policy $\pi_c$ of \Cref{assumption:1} and the deterministic optimal policy $\pi^\star$ of \Cref{eqn:discounted_optimal_policy} are optimized using CCMs based on a pretrained dynamics model $\dot{\hat{x}} = \hat{f}_\xi(x) + \hat{B}_\zeta(x) u$. Given calibration trajectories $\{x_j(t),u_j(t)\}_{j=1}^{N_{\text{cal}}}$ not used for pretraining $\dot{\hat{x}}$, we define the nonconformity score quantifying the model mismatch as follows:
    \begin{align}
        \label{eq_score}
        \overline{\mf{d}}_j \coloneq \sup_{t\geq t_0} \|\mf{d}(x_j(t),u_j(t))\|, 
        \quad j=1,\cdots,N_{\text{cal}}.
    \end{align}
    If these scores and the score computed with the contracting policy $\pi_c$ are exchangeable and $\pi^\star$ achieves a lower cumulative tracking error than $\pi_c$, then a test trajectory executed using $\pi^\star$ satisfies the following bound on the time-averaged $\ell^2$-norm:
    \begin{align}
        \label{eq_robust_bound}
        \mb{P}\left[\lim_{\kappa\to\infty}\frac{\|(\delta x)_\kappa\|_{\ell^2_M}}{\kappa} \leq \frac{\sqrt{\overline{m}}\mf{Q}_{1-\varepsilon}}{\lambda}\right] \geq 1-\varepsilon.
    \end{align}
    Here, $\varepsilon\in(0,1)$ is a user-specified risk tolerance level, $\mf{Q}_{1-\varepsilon}$ is the $(1-\varepsilon)$-th quantile of the empirical distribution of $\{\overline{\mf{d}}_j\}_{j=1}^{N_{\text{cal}}}\cup\infty$, and $x$ is the state of \Cref{eqn:control_affine_system} under the policy $\pi^\star$. Note that the probability measure $\mb{P}$ is taken over the randomness of the calibration trajectories and the test trajectory~\cite{tingwei_conformal}.
\end{theorem}

\section{Experimental Results: Simulation} \label{sec:experiments}
We conducted experiments in simulated environments to evaluate CARL’s performance and robustness, implied in \Cref{thm:asymptotic_stability,thm_robustness}, relative to a set of relevant baselines.

\subsection{Simulated Environments}\label{subsec:simulated_environments}

\begin{figure}
    \centering
    \includegraphics[width=0.98\linewidth]{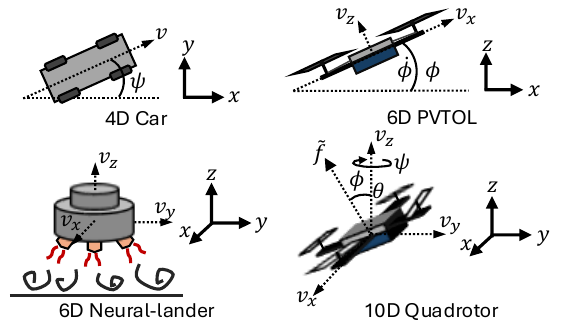}
    \caption{Notional illustrations of the robots used in our simulated experiments and their underlying dynamics.}
    \label{fig:simulation_robots}
\end{figure}

We considered four widely used robot models for path-tracking experiments \cite{sun2021learning, liu2020robust, singh2021learning, singh2023robust}: 4D Car, 6D PVTOL, 6D Neural-lander, and 10D Quadrotor. 
We visualize these models in \Cref{fig:simulation_robots} and define the corresponding dynamics below.
These dynamics were only used to define our simulation environments---we assume they are unknown for all considered algorithms.

\subsubsection{4D Car}
The state of the car is defined as $x \coloneq (p_x, p_y, v, \psi)$, where $p_x$ and $p_y$ define the 2D position, $v$ is the linear speed along the yaw angle, and $\psi$ is the yaw angle. 
The control input is defined as $u \coloneq (a, \omega)$, where $a$ is the linear acceleration and $\omega$ is the angular velocity. 
The dynamics are expressed as follows:
\begin{equation}
    f(x) =
    \begin{bmatrix}
    v \cos(\psi) \\
    v \sin(\psi) \\
    0 \\
    0 
    \end{bmatrix}, \quad
    B(x) =
    \begin{bmatrix}
    0 & 0 \\
    0 & 0 \\
    1 & 0 \\ 
    0 & 1 \\ 
    \end{bmatrix}.
\end{equation}
This design is a relaxation of the classic Dubins car, as it allows the linear velocity to vary (rather than being fixed). 
This is the environment with the lowest dimensionality and dynamic complexity that we consider.

\subsubsection{6D PVTOL}
We adopted the PVTOL model from \cite{sun2021learning, singh2021learning, singh2023robust}.
The state of the PVTOL is defined as $x \coloneq (p_x, p_z, \phi, v_x, v_z, \dot{\phi})$, where $p_x$ and $p_z$ define the 2D position, $\phi$ is the roll angle, $v_x$ and $v_z$ are the linear velocities, and $\dot{\phi}$ is the roll angle rate. 
The control input is defined as $u\coloneq (\tilde f_1, \tilde f_2)$, where $\tilde f_1$ and $\tilde f_2$ are the thrust of two rotors that controls the $v_z$ and $\dot{\phi}$. 
The dynamics are expressed as follows:
\begin{equation}
    f(x) =
    \begin{bmatrix}
    v_x \cos(\phi) - v_z \sin(\phi) \\
    v_x \sin(\phi) + v_z \cos(\phi) \\
    \dot \phi \\
    v_z \dot \phi - g \sin(\phi) \\
    -v_x \dot \phi -  \cos(\phi) \\
    0
    \end{bmatrix}, \;
    B(x) =
    \begin{bmatrix}
    0 & 0 \\
    0 & 0 \\
    0 & 0 \\
    0 & 0 \\
    1/\tilde{m} & 1/ \tilde{m} \\ 
    l/\mc{I} & -l/\mc{I} \\ 
    \end{bmatrix},
\end{equation}
where $g$ is the gravitational acceleration, $\tilde{m}$ is the mass, $l$ is the rotor arm length, and $\mc{I}$ is the moment of inertia.

\subsubsection{6D Neural-lander}
We adopted the Neural-lander model from \cite{liu2020robust, sun2021learning}.
The state of the Neural-lander is defined as $x \coloneq (p_x, p_y, p_z, v_x, v_y, v_z)$, where $p_x$, $p_y$, and $p_z$ define the 3D position, and $v_x$, $v_y$, and $v_z$ are the 3D linear velocities. 
The control input is defined as the 3D linear accelerations as $u \coloneq (a_x, a_y, a_z)$. 
The dynamics are expressed as follows:
\begin{equation}
    f(x) =
    \begin{bmatrix}
    v_x \\
    v_y \\
    v_z \\
    F_1/\tilde m \\
    F_2 / \tilde m \\
    F_3 /\tilde m - g
    \end{bmatrix}, \quad
    B(x) =
    \begin{bmatrix}
    0 & 0 & 0 \\
    0 & 0 & 0 \\
    0 & 0 & 0 \\
    1 & 0 & 0 \\
    0 & 1 & 0 \\ 
    0 & 0 & 1 \\ 
    \end{bmatrix},
\end{equation}
where $F_i \coloneq F_i(p_z, v_x, v_y, v_z)$ are neural network functions that were trained to predict the ground effect for each dimension.
\cite{liu2020robust} developed this environment to consider the ground effect of a flying robot when landing. 

\subsubsection{10D Quadrotor}
We adopted the Quadrotor model from \cite{sun2021learning, singh2023robust}.
The state of the Quadrotor is defined as $x \coloneq (p_x, p_y, p_z, v_x, v_y, v_z, \tilde f, \phi, \theta, \psi)$, where $p_x$, $p_y$, and $p_z$ define the 3D position, $v_x$, $v_y$, and $v_z$ define the 3D linear velocities, $\tilde f$ is the thrust, and $\phi$, $\theta$, and $\psi$ are the attitude. 
The control input is defined as the rate of change of thrust and attitude as $u \coloneq (\dot{\tilde{f}}, \dot \phi, \dot \theta, \dot \psi)$. 
The dynamics are expressed as follows:
\begin{equation}
    f(x) =
    \begin{bmatrix}
    v_x \\
    v_y \\
    v_z \\
    \tilde f \sin(\psi) \\
    -\tilde f \cos(\psi) \sin(\phi) \\
    \tilde f \cos(\psi)\cos(\phi) - g\\ 
    0 \\
    0 \\
    0 \\
    0 \\
    \end{bmatrix}, \;
    B(x) =
    \begin{bmatrix}
    0 & 0 & 0 & 0 \\
    0 & 0 & 0 & 0 \\
    0 & 0 & 0 & 0 \\
    0 & 0 & 0 & 0 \\
    0 & 0 & 0 & 0 \\
    0 & 0 & 0 & 0 \\
    1 & 0 & 0 & 0 \\
    0 & 1 & 0 & 0 \\ 
    0 & 0 & 1 & 0 \\
    0 & 0 & 0 & 1 \\
    \end{bmatrix}.
\end{equation}

\subsection{Baseline Algorithms}
We considered the following baseline algorithms for comparison:
\begin{enumerate}
    \item Certified control using contraction metric (C3M) \cite{sun2021learning},
    \item Proximal policy optimization (PPO) \cite{schulman2017proximal},
    \item State-dependent LQR (SD-LQR) \cite{richards2023learning},
    \item Standard LQR \cite{kalman1960contributions}.
\end{enumerate}

C3M is a learning-based algorithm that simultaneously identifies CCMs and a corresponding contracting policy using two neural networks. 
PPO is a widely used actor-critic algorithm that CARL is based on. 
SD-LQR builds on LQR by introducing exact nonlinear error dynamics using state-dependent coefficients; it solves the state-dependent Riccati equation, which practically results in a larger region of attraction than LQR \cite{cimen2012survey,richards2023learning}.

The RL algorithms (CARL and PPO) take $s_t = (x_t, x_{t:t+N-1}^*, u_t^*)$ as input to generate controls, while the control-theoretic baselines (C3M, SD-LQR, and LQR) take $s_t = (x_t, x_t^*, u_t^*)$ since their control laws inherently embed the system dynamics for path-tracking.
The baseline PPO is optimized using the reward function in \Cref{eqn:our_reward_function} with entropy regularization and CCMs replaced by the identity matrix (i.e., $M = I$).
We list the key hyperparameters used for CARL and the baselines in \Cref{tab:hyperparameters}.

We adopted $\lambda$, $\overline{m}$, and $\underline{m}$ from Sun et al. ~\cite{sun2021learning} which considers similar dynamical systems, and tuned $\beta_\mathcal{M}$, $\beta_\pi$, and the target KL divergence using a grid search.

\begin{table}[t]
    \centering
    \caption{List of key hyperparameters used in the experiments. 
    The CCM parameters $\lambda$ and $\overline{m}, \underline{m}$ govern the speed and transient overshoot of the contraction, respectively, while the RL parameters are standard configurations for general policy optimization.
    }
    \label{tab:hyperparameters}
    \begin{tabular}{lcc}
        \toprule
        \textbf{Parameter} & \textbf{Symbol} & \textbf{Value} \\
        \midrule
        \multicolumn{3}{l}{\textbf{CCM parameters}} \\
        Target contraction rate & $\lambda$ & 0.5 \\
        CCM upper bound & $\overline{m}$ & 10.0 \\
        CCM lower bound & $\underline{m}$ & 0.1 \\
        Entropy-regularization & $\beta_\mc{M}$ & 0.01 \\
        \midrule
        \multicolumn{3}{l}{\textbf{RL parameters}} \\
        Discount factor & $\gamma$ & 0.99 \\
        Generalized advantage estimator & - & 0.95 \\
        Target KL divergence & - & 0.01 \\
        Entropy-regularization & $\beta_\pi$ & 0.001 \\
        \bottomrule
    \end{tabular}
\end{table}

\subsection{Experiment Details}\label{subsec:simul_exp_detail}
\subsubsection{Reference trajectory generation}\label{subsubsec:simul_reference_traj}
We generated reference trajectories following Sun et al. \cite{sun2021learning}. 
Specifically, we defined a compact set of initial reference states $\mathcal{X}_0^{*} \subset \mathcal{X}$, from which the initial state of the reference trajectory is uniformly sampled, i.e., $x^*(0) \sim \mathrm{Unif}(\mathcal{X}_0^*)$.
A reference control $ u^*(t) $ was then constructed as a convex combination of $N_\text{ref}$ sinusoidal functions as follows:
\begin{equation}\label{eqn:reference_control}
    u^*(t) = \sum_{i=1}^{N_\text{ref}} w_i \sin\left(2\pi \varrho_i \frac{t}{T_{\max}}\right),
\end{equation}
where $ \varrho_i $ are fixed frequencies, $ w_i $ are normalized weights that sum to $1$, and $ T_{\max} $ is the maximum time horizon.
We set $N_{\text{ref}} = 10$, defined the frequencies as $\varrho_i = i$, and sampled weights $w_i \sim \mathrm{Unif}(0, 1)$, which were subsequently normalized.
We ensured the trajectory remains within the predefined compact set $\mathcal{X}$ by truncating the episode when the Cartesian coordinates exceed the bounds and clipping the remaining components. 
This process may result in reference trajectories of varying time lengths.

\subsubsection{Pretrained dynamics model}
\label{sec:dynamics_model}
\begin{figure}[t]
    \centering
    \includegraphics[width=0.99\linewidth]{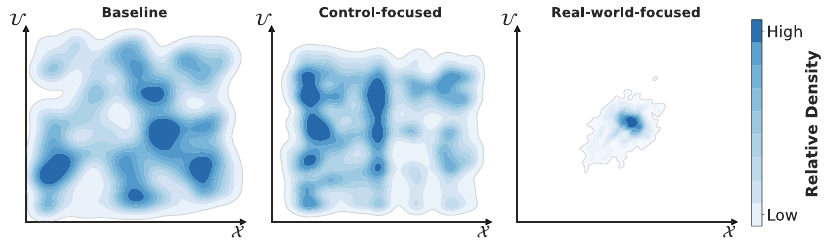}
    \caption{Notional density plots of the sampled state ($\mathcal{X}$) and control ($\mathcal{U}$) spaces for pretraining dynamics models. The baseline dataset shows a broad, uniform distribution in the state space with limited coverage in the control space. The real-world dataset is constrained to a highly specific region. The control-focused dataset spans the control space extensively but remains limited in the state space.}
    \label{fig:nominal_data}
\end{figure}

We constructed three datasets of differing coverage of the state and control spaces. 
Our baseline data contained $N_{\text{data}} = 100,000$ homogeneous samples of $(\dot{x}, x, u)$ data points, generated by uniformly sampling $N_\text{data}/N_{\text{ctr}}$ states from $\mathcal{X}$, pairing each with $N_{\text{ctr}} = 1$ control inputs uniformly sampled from $\mathcal{U}$, and computing corresponding dynamics $\dot x$ using the corresponding nominal dynamics model in \Cref{subsec:simulated_environments}. 
This procedure follows Richards et al. \cite{richards2023learning}.
We also constructed control-focused data generated using the same procedure but with $N_{\text{ctr}}=3$ to provide better coverage of the control space, but with less coverage of the state space.
Finally, we constructed real-world-focused data that better reflects real-world constraints, containing only $10,000$ samples of reference trajectories generated by adding Gaussian noise to the reference controls.
We provide notional illustrations of these datasets in \Cref{fig:nominal_data}.

We used these datasets to pretrain models of the system dynamics, $ \hat{f}_{\xi} $ and $\hat{B}_{\zeta}$ (see \Cref{sec:pretraining-dynamics-model}). Each model was represented by a feed-forward neural network with hidden layers of size (256, 256) and Tanh activation functions. 
We refer to the dynamics models trained with baseline, control-focused, and real-world-focused data as the baseline, control-focused, and real-world-focused dynamics models, respectively.

\subsubsection{CMG network}
Following Sun et al. \cite{sun2021learning}, we modeled the CMG using a feed-forward neural network with hidden layers of size (256, 256) and Tanh activation functions. 
However, unlike \cite{sun2021learning}, we introduced dual output layers to generate the mean and variance for parameterizing univariate Gaussian distributions over CCMs.

\subsubsection{Policy network}
Following Sun et al. \cite{sun2021learning}, we modeled the policy for CARL, C3M, and PPO using two feed-forward neural networks with hidden layers of size (128) and Tanh activation functions, parameterized by $\theta = \{\theta_{1}, \theta_{2} \}$, and defined controls as follows:
\begin{equation}\label{eqn:control_synthesis_law}
    u_t = u_t^* + w_2^\top \cdot \tanh(w_1 \cdot (x_t - x_t^*)),
\end{equation}
where $u^*$ is the reference control used to generate the next reference state, $w_1=\theta_{1}(x_t, x_{t:t+N-1}^*) \in \mb{R}^{n\times n}$, and $w_2=\theta_{2}(x_t, x_{t:t+N-1}^*) \in \mb{R}^{n \times m}$.
C3M synthesizes controls using \Cref{eqn:control_synthesis_law} with $w_1$ and $w_2$ taking ($x_t, x_t^*$).

\subsubsection{Evaluation metric}
We first define the normalized tracking error for time $t_k = k \Delta t$, with discrete time interval $\Delta t$,  as
\begin{equation}\label{eqn:normalized_tracking_error}
    \| \mathrm e (t_k) \|_2 \coloneq \frac{\|x(k\Delta t) - x^*(k\Delta t)\|_2}{\|x(0) - x^*(0)\|_2}.
\end{equation}
Then, similar to \cite{sun2021learning,richards2023learning}, we introduce a metric for path-tracking performance under varying lengths of reference trajectories via the modified area under the curve (mAUC) of the normalized tracking error as follows:
\begin{equation}\label{eqn:mAUC}
    \text{mAUC} \coloneq \frac{L_{\max}}{L} \sum_{k=0}^{L} \| \mathrm e (t_k) \|_2,
\end{equation}
where $L_{\max} = T_{\max} / \Delta t$, $L = T / \Delta t$, and $T$ is the time taken for the executed trajectory. 
We use mAUC as a primary path-tracking performance indicator, as it captures the cumulative tracking error for varying lengths of reference trajectories.

We evaluated the path-tracking performance of a policy by considering 10 different reference trajectories. For each reference trajectory, we performed rollouts from 10 different initial states near the initial state of that reference trajectory. We then calculated the mean and $95\%$ confidence interval of mAUC over those 100 trajectories.

\subsection{Simulation Results}
We report mean and $95\%$ confidence intervals of mAUC and the mean per-step inference time---measured on an Intel i7-1165G7 (2.80GHz) CPU with 16GB of 4266MHz LPDDR4X SDRAM---of each algorithm over 5 random seeds in \Cref{table:mauc_report} and depict their corresponding normalized error curves in \Cref{fig:simulation_plot}.
Additionally, we visualize the path-tracking results in Appendix \ref{app:results} (\Cref{fig:simulation_trajectory}).
We highlight four key observations.

\begin{table*}[t]
\centering
\footnotesize
\setlength{\tabcolsep}{5pt}  
\caption{\textbf{Evaluation results with baseline dynamics models.} Mean and 95\% confidence intervals of mAUC over 5 seeds are reported with mean per-step inference time in milliseconds (ms). The best result for each environment is bolded.}
\label{table:mauc_report}
\begin{tabular}{@{}l cccc | cccc@{}}
    \toprule
    & \multicolumn{4}{c}{\textbf{mAUC (lower is better)}} & \multicolumn{4}{c}{\textbf{Per-step Inference Time (ms)}} \\
    \cmidrule(lr){2-5} \cmidrule(lr){6-9}
    \textbf{Method} & Car & PVTOL & Neural-lander & Quadrotor & Car & PVTOL & Neural-lander & Quadrotor \\
    \midrule
    CARL (Ours) & $1.07\pm0.05$ & $\mathbf{7.49\pm0.84}$ & $\mathbf{2.47\pm0.42}$ & $\mathbf{5.55\pm1.36}$ & $\mathbf{0.1}$ & $\mathbf{0.1}$ & $\mathbf{0.1}$ & $\mathbf{0.1}$ \\
    C3M \cite{sun2021learning} & $\mathbf{1.06\pm0.07}$ & $29.6\pm6.4$ & $2.63\pm0.44$ & $6.97\pm1.72$ & $\mathbf{0.1}$ & $\mathbf{0.1}$ & $\mathbf{0.1}$ & $\mathbf{0.1}$ \\
    PPO \cite{schulman2017proximal} & $1.56\pm0.18$ & $12.7\pm3.8$ & $2.64\pm0.48$ & $5.84\pm1.13$ & $\mathbf{0.1}$ & $\mathbf{0.1}$ & $\mathbf{0.1}$ & $\mathbf{0.1}$ \\
    SD-LQR \cite{richards2023learning} & $7.88\pm0.19$ & $16.7\pm1.21$ & $5.67\pm0.36$ & $5.73\pm1.03$ & $1$ & $10$ & $10$ & $20$ \\
    LQR \cite{kalman1960contributions} & $9.6\pm0.22$ & $17.7\pm0.8$ & $5.7\pm0.36$ & $5.79\pm0.98$ & $4$ & $10$ & $10$ & $30$ \\
    \bottomrule
\end{tabular}
\end{table*}

\begin{figure*}[t]
    \centering
    \includegraphics[width=0.99\linewidth]{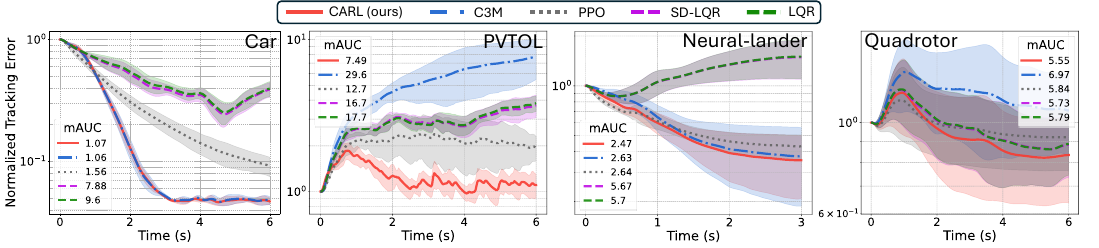}
    \caption{We plot the mean and $95\%$ confidence intervals of the normalized tracking error, $\|\mathrm{e}(t)\|_2$, for CARL and relevant baselines across 5 random seeds.}
    \label{fig:simulation_plot}
\end{figure*}

\subsubsection{Performance of CARL} As shown in \Cref{table:mauc_report}, CARL achieves the best performance in the PVTOL, Neural-lander, and Quadrotor environments, and demonstrates comparable performance to C3M in Car (the simplest environment).
CARL shows especially strong performance in PVTOL, where it achieves a $41\%$ reduction in mAUC over the next-best baseline.
We attribute these results to the cumulative minimization of both transient and steady-state errors in CARL.
Additionally, CARL consistently outperforms PPO, highlighting the advantage of our CCM-conditioned reward function. 
Finally, LQR-based methods perform poorly because LQR relies on linearized dynamics, and neither LQR nor SD-LQR guarantees stabilization of the closed-loop error dynamics for nonlinear systems \cite{richards2023learning}.

\subsubsection{Practicality of CARL} 
\Cref{table:mauc_report} also shows that CARL, C3M, and PPO achieve low (0.1 ms) per-step inference time of controls regardless of system dimensionality, due to the use of feed-forward networks.
In comparison, LQR-based methods have nearly $100\times$ longer inference times for solving the Riccati equation at each time step.
Moreover, inference time grows with problem dimensionality, indicating that these methods do not scale well. Such scaling is important, given that most robotic systems, such as drones and mobile robots, operate with limited computational capacity in high-dimensional environments.

\begin{table}[t]
\centering
\scriptsize 
\setlength{\tabcolsep}{5pt}  
\caption{\textbf{Evaluation results with dynamics models of varying coverage.} Mean and 95\% confidence intervals of mAUC over 5 seeds are reported with the best results bolded. PPO is omitted as it does not use a dynamics model.}
\label{table:mauc_ablation}
\begin{tabular}{@{}l cccc@{}}
    \toprule
    & \multicolumn{4}{c}{\textbf{mAUC (lower is better)}} \\
    \cmidrule(lr){2-5}
    \textbf{Method} & Car & PVTOL & Neural-lander & Quadrotor \\
    \midrule\midrule
    \multicolumn{5}{c}{\textbf{Control-focused dynamics}} \\
    \midrule
    CARL (Ours) & $\mathbf{1.05\pm0.06}$ & $\mathbf{7.12\pm0.75}$ & $\mathbf{2.33\pm0.39}$ & $\mathbf{5.42\pm1.21}$ \\
    C3M \cite{sun2021learning} & $1.06\pm0.08$ & $19.6\pm11$ & $2.44\pm0.39$ & $5.68\pm1.27$ \\
    SD-LQR \cite{richards2023learning} & $7.51\pm2.81$ & $11.0\pm1.04$ & $5.68\pm2.31$ & $6.27\pm1.49$ \\
    LQR \cite{kalman1960contributions} & $17.5\pm7.38$ & $10.8\pm1.11$ & $5.72\pm2.13$ & $6.48\pm1.41$ \\
    \midrule\midrule
    \multicolumn{5}{c}{\textbf{Real-world-focused dynamics}} \\
    \midrule
    CARL (Ours) & $\mathbf{1.47\pm0.11}$ & $\mathbf{7.87\pm1.14}$ & $\mathbf{2.53\pm0.83}$ & $\mathbf{5.75\pm1.54}$ \\
    C3M \cite{sun2021learning} & $10.8\pm1.8$ & $33.2\pm12.9$ & $3.88\pm1.13$ & $5.96\pm1.63$ \\
    SD-LQR \cite{richards2023learning} & $15.4\pm5.3$ & $26.7\pm7.5$ & $9.43\pm1.0$ & $6.65\pm1.1$ \\
    LQR \cite{kalman1960contributions} & $16.1\pm6.4$ & $32.5\pm11.2$ & $9.49\pm0.87$ & $6.34\pm0.84$ \\
    \bottomrule
\end{tabular}
\end{table}

\subsubsection{Robustness of CARL} 
We also studied the robustness of CARL to the varying coverage of data used for learning the dynamics model, by considering the control-focused and real-world-focused dynamics models introduced in \Cref{sec:dynamics_model}.
We considered the control-focused dynamics models to understand the impact of emphasizing better coverage of the control space, rather than the state space.
We considered the real-world-focused dynamics models to consider a more realistic setting where we cannot collect data uniformly.
We do not show results for PPO because it does not use a dynamics model.

As shown in \Cref{table:mauc_ablation}, CARL outperforms all baselines across all environments when using the control-focused and real-world-focused dynamics models, often by a significant margin. 
CARL likely outperforms C3M in these settings because it does not directly rely on the approximate dynamics in policy optimization.
Specifically, while C3M synthesizes the policy using the approximate CCMs (which are biased metrics due to approximate dynamics), CARL instead incorporates them implicitly through the reward function, where the maximum reward is achieved when the system contracts, regardless of approximation errors in the pretrained dynamics during simulation-based training.

The lack of robustness for LQR-based methods is likely due to their need for linearized dynamics (LQR) and additional approximation of state-dependent coefficients of the dynamics (i.e., $f(x) = A(x)x$) from the approximate dynamics model (SD-LQR), making them more susceptible to model approximation errors.

We also observe that CARL, when evaluated under real-world-focused dynamics, performs better than PPO. 
This suggests that CARL has the potential to outperform PPO in real-world settings, which is consistent with our robot experimental results, as we discuss in \Cref{sec:experiment_ground_robot} and \Cref{sec:experiment_aerial_robot}.

\subsubsection{Ablation study of CARL} 
Our ablation study compares the performance of CARL—which employs a stochastic, entropy-regularized CMG—against two variants: a deterministic CMG and a stochastic CMG without entropy regularization.  
\Cref{table:ablation_transposed} reports the mAUC obtained by CARL under each CMG variant.
We see that none of the CMG variants yielded better performance (lower mAUC) than the default configuration, except for the stochastic CMG without entropy regularization in Quadrotor.
We expect the improved performance of the stochastic CMG without entropy regularization in Quadrotor may be due to tuning of the $\beta_\mc{M}$ hyperparameter, which regulates the strength of the entropy regularization.

\begin{table}[t]
\centering
\scriptsize
\setlength{\tabcolsep}{2pt} 
\caption{\textbf{Ablation results for CARL with baseline dynamics model}. Mean and 95\% confidence intervals of mAUC over 5 seeds are reported with the results that are better than one in \Cref{table:mauc_report} bolded.}
\label{table:ablation_transposed}
\begin{tabular}{@{}l ccc@{}}
    \toprule
    \multicolumn{4}{c}{\textbf{CARL ablation: mAUC (lower is better)}} \\
    \cmidrule(lr){2-4}
    \textbf{Environment} & \textbf{Default} & \textbf{Stochastic CMG w/o entropy} & \textbf{Deterministic CMG} \\
    \midrule\midrule
    Car & $\mathbf{1.07 \pm 0.05}$ & $1.08 \pm 0.07$ & $1.08\pm 0.08$ \\
    PVTOL & $\mathbf{7.49 \pm 0.84}$ & $7.91\pm 1.11$ & $8.79\pm2.21$ \\
    Neural-lander & $\mathbf{2.47 \pm 0.42}$ & $2.54\pm0.58$ & $2.51\pm0.38$ \\
    Quadrotor & $5.55 \pm 1.36$ & $\mathbf{5.29\pm 1.55}$ & $5.64\pm1.45$ \\
    \bottomrule
\end{tabular}
\end{table}

\section{Real-World Experiment: TurtleBot3 Burger}\label{sec:experiment_ground_robot}
\begin{figure*}[t!]
    \centering
    \includegraphics[width=0.99\linewidth]{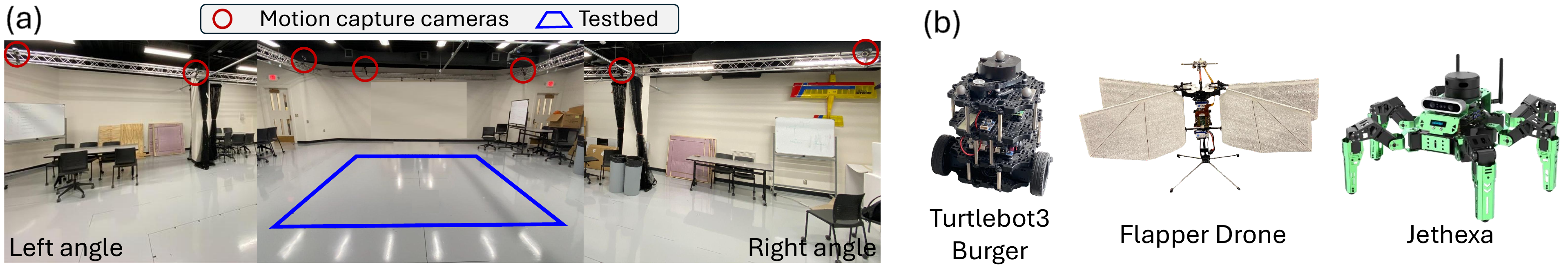}
    \caption{(a) Our facility for real-world robot experiments, which include a $12 \times 12$ ft testbed. 
    \href{https://www.qualisys.com/}{Qualisys} motion capture cameras are denoted by a red circle. (b) We consider wheeled, aerial, and legged robots.}
    \label{fig:experiment_facility}
\end{figure*}

We evaluated CARL in three challenging sim-to-real settings.
We used the facility depicted in \Cref{fig:experiment_facility}a, which has eight \href{https://www.qualisys.com/}{Qualisys} motion capture cameras to track the position and orientation of the robots. 
We considered wheeled, aerial, and legged robotic platforms (depicted in \Cref{fig:experiment_facility}b).
This section focuses on the TurtleBot3 Burger experiment.
We discuss the Flapper Drone experiment in \Cref{sec:experiment_aerial_robot} and the Jethexa experiment in \Cref{sec:experiment_legged_robot}.
A video of all experiments is available at this \href{https://youtu.be/sOJ4hulbop0}{link}.

\begin{figure*}[t]
    \centering
    \includegraphics[width=0.95\linewidth]{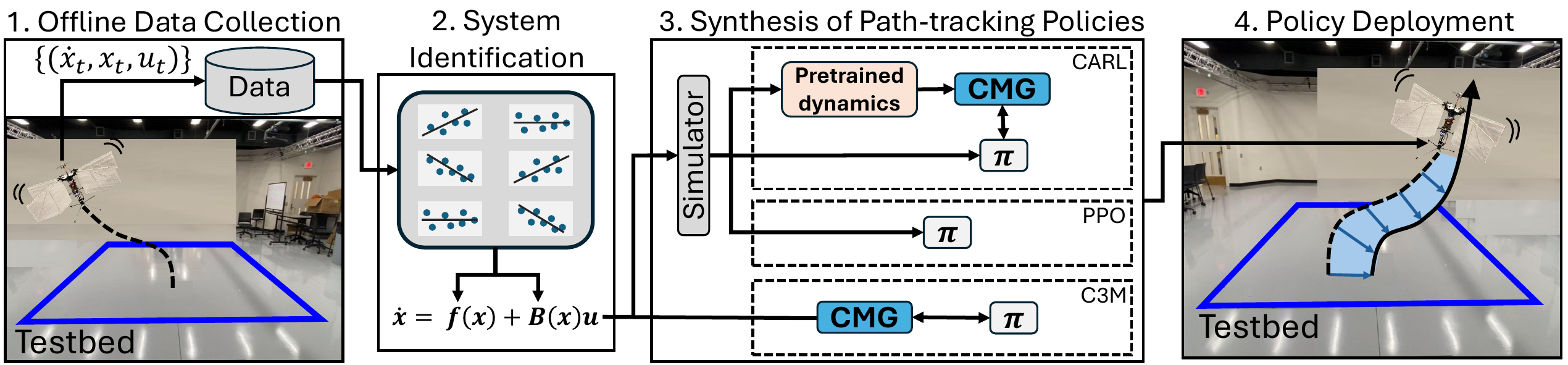}
    \caption{Our real-world experimental pipeline consists of the following stages: 1) collecting offline data, 2) system identification, 3) policy synthesis, and 4) policy deployment.
    }
    \label{fig:cac_high_level}
\end{figure*}

Our experiment pipeline is shown in \Cref{fig:cac_high_level} and consists of four stages: 1) collecting offline data, 2) conducting system identification, 3) synthesizing a path-tracking policy, and 4) deploying the policy in the real-world.
We only considered C3M and PPO as baselines in these real-world experiments, due to the high per-step inference time of LQR-based algorithms.
For CARL, we used the identified system model to create a simulator for pretraining a dynamics model and policy optimization.
We gave C3M direct access to the identified system model (i.e., we do not use a pretrained dynamics model).
Similar to CARL, PPO used the identified system model to create a simulator for policy optimization.

We did not implement any specific sim-to-real techniques, such as domain randomization \cite{tobin2017domain} or mini-max game for worst-case optimization \cite{smirnova2019distributionally}.
We did this to focus our evaluation on the inherent robustness of our considered algorithms to errors from the dynamics mismatch between simulation and the real-world and model approximation errors in formulating CCM conditions.
In principle, CARL can be implemented with existing sim-to-real techniques to further reduce the sim-to-real gap.
Unless otherwise specified, the experimental setup follows that of our simulated experiments, as discussed in \Cref{subsec:simul_exp_detail}.

\subsection{TurtleBot3 Burger Experiment Details}\label{subsubsec:turtlebot_exp_details}

\subsubsection{Offline data collection}\label{subsubsec:turtlebot_offline_data_collection}
We constructed an offline dataset using 26 reference trajectories.
Following the simulated real-world-focused data generation process described in \Cref{subsec:simul_exp_detail}, each reference trajectory was generated by initializing the robot and executing reference controls with added Gaussian noise from that state.
The motion capture system recorded the resulting sequence of positions and attitudes of the robot, which we used as reference trajectories.
We then used a second-order finite difference method to compute the dynamics $\dot x$.
We used 20 reference trajectories for training, five for validation, and one for testing (evaluation). 
Training trajectories were used for system identification and policy optimization, while validation trajectories were used to identify the best-performing checkpoint for deployment. 
The best-performing policy was then deployed using the test trajectory as a reference trajectory to evaluate path-tracking performance.

\subsubsection{System identification}\label{subsubsec:turtlebot_system_id}
We modeled the dynamics of the TurtleBot3 Burger as follows:
\begin{equation}\label{eqn:turtlebot_nominal_dynamics}
    f(x) =
    \begin{bmatrix}
    0 \\
    0 \\
    0
    \end{bmatrix}, \quad
    B(x) =
    \begin{bmatrix}
    c_1 \cos(\psi) & 0 \\
    c_2 \sin(\psi) & 0 \\
    0 & c_3
    \end{bmatrix},
\end{equation}
where the state is defined as $x \coloneq (p_x, p_y, \psi)$, with $p_x$ and $p_y$ denoting the 2D position and $\psi$ denoting the yaw angle.
The control input is defined as $u \coloneq (v, \omega)$, representing linear and angular velocities, respectively. 
We identified the empirical coefficients using least-squares regression with 20 training trajectories as follows:
\begin{equation}\label{eqn:least-squares}
    c_i = \argmin_{c \in \mathbb{R}} \sum_{j=1}^{20} \sum_{t=1}^{L_{\max}} \left( \dot{x}_{i, t}^{(j)} - c \cdot \dot{\hat{x}}_{i, t}^{(j)} \right)^2, \quad \forall i \in \{1, 2, 3\},
\end{equation}
where $j$ is the trajectory index and $t$ is the time step. 
Here, $\dot{x}_{i, t}^{(j)}$ denotes the observed dynamics from the motion capture system for the $i$-th dimension of the state, and $\dot{\hat{x}}_{i, t}^{(j)}$ denotes the corresponding nominal dynamics term (i.e., the $i$-th row of \Cref{eqn:turtlebot_nominal_dynamics} excluding the empirical coefficient $c_i$).
We show the identified empirical coefficients in \Cref{tab:tuetlrbot_empirical_coefficients} and visualizations of the least-squares fits in Appendix \ref{app:results} (\Cref{fig:lstsq}a).
These coefficients reflect uncertainties such as friction and actuator delays (e.g., motor response and command latency).
\begin{table}[t]
    \centering
    \caption{Identified empirical coefficients for the TurtleBot3 Burger dynamics.}
    \label{tab:tuetlrbot_empirical_coefficients}
    \begin{tabular}{@{}cc@{}}
        \toprule
        \textbf{Symbol} & \textbf{Value} \\
        \midrule
        $c_1$ & 0.8183 \\
        $c_2$ & 0.8179 \\
        $c_3$ & 0.6767 \\
        \bottomrule
    \end{tabular}
\end{table}

\subsubsection{Synthesis of path-tracking policies}\label{subsubsec:turtlebot_policy_optimization}
For CARL and PPO, we optimized the policy by interacting with a simulator that integrates the identified dynamics from \Cref{subsubsec:turtlebot_system_id} using the Euler method. 
For C3M, we used the same identified dynamics to directly synthesize the contracting policy.

\subsubsection{Policy deployment}
We executed policies on the robot at 10 Hz on a centralized computer for 20 seconds. 
We constructed the policy input (state) by using real-time 3D pose data from the motion capture cameras.
We passed the controls output by the policy to the onboard velocity PID controller, which implements corresponding linear and angular velocities.

\subsubsection{Evaluation metric}
\label{sec:ground-evaluation-metric}
We evaluated path-tracking performance using mAUC, as defined in \Cref{eqn:mAUC}, by considering the held out test reference trajectory. 
We randomly initialized five different initial states for the robot near the initial state of this test trajectory, and rollout one trajectory for each initial state.
We then calculated the mean and $95\%$ confidence interval of mAUC over those five trajectories.

\subsection{TurtleBot3 Burger Results}\label{subsec:turtlebot_exp_details}
\begin{figure*}[t!]
    \centering
    \includegraphics[width=0.8\linewidth]{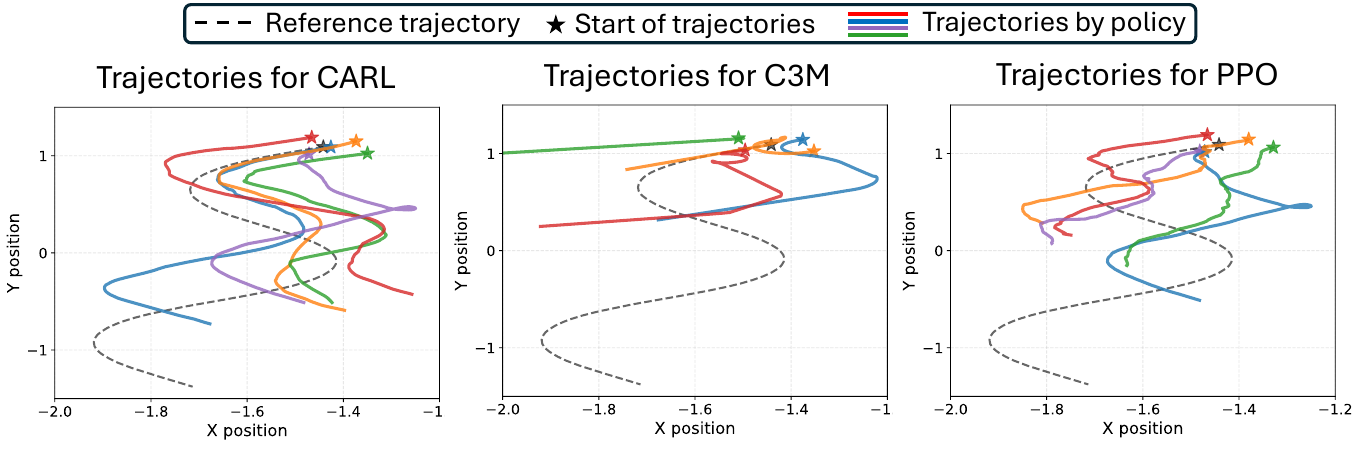}
    \caption{We demonstrated our algorithm (CARL) on a TurtleBot3 Burger along with key baselines. The robot was initialized at five distinct initial positions to track the same reference trajectory across all algorithms. 
    }
    \label{fig:robot_trajectory}
\end{figure*}

\begin{figure}[t!]
    \centering
    \includegraphics[width=0.9\linewidth]{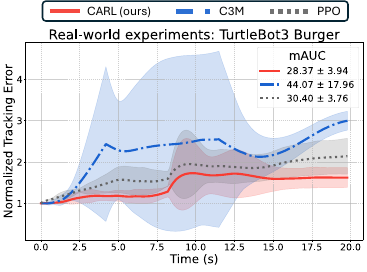}
    \caption{Mean and 95\% confidence intervals for the normalized tracking error over 5 seeds from real-world experiments. The higher mAUC observed, compared to that in \Cref{table:mauc_report}, is attributed to the longer horizon length and increased uncertainty from the sim-to-real transfer.}
    \label{fig:robot_tracking_result}
\end{figure}
\Cref{fig:robot_trajectory} shows the reference trajectory (i.e., test trajectory) and executed trajectories from deploying trained policies on a TurtleBot3 Burger with five different initial states for CARL, C3M, and PPO.
\Cref{fig:robot_tracking_result} shows the corresponding normalized tracking errors for these trajectories, with the mean and 95\% confidence intervals.
We see that CARL outperforms C3M and PPO, where C3M completely fails to track the path, and PPO exhibits instability during path-tracking with noticeable jittering. 
In contrast, CARL maintains the most reliable trajectory tracking with the lowest mAUC, demonstrating its robustness in a real-world setting.

\section{Real-World Experiment: Flapper Drone}\label{sec:experiment_aerial_robot}
We also evaluated CARL using the \href{https://flapper-drones.eu/}{Flapper Drone} depicted in \Cref{fig:experiment_facility}b, within the same facility shown in \Cref{fig:experiment_facility}a.
We followed the same experimental pipeline shown in \Cref{fig:cac_high_level}.

\subsection{Flapper Drone Experimental Details}

\subsubsection{Cascaded control approach}
\label{sec:cascaded-control-approach}
In contrast to the TurtleBot3 robot, the Flapper Drone exhibits challenging instabilities due to its non-rigid wings and nonlinear thrust-voltage dependency, rendering direct control using an on-board velocity-based (low-level) PID controller ineffective. 
Consequently, we implemented a cascaded control architecture where the policy's low-level commands (e.g., thrust and attitude rate) were first converted into the desired next state (3D position and yaw angle) using the identified dynamics model that we discuss in \Cref{subsubsec:flapper_system_id}. 
These computed targets were then tracked by the robot's onboard position-based (high-level) PID controller, which internally generates low-level commands for the velocity-based PID loop.
Our experiments show this cascaded architecture to be much more reliable than direct control for all algorithms considered.

\subsubsection{Offline data collection}
Since we employ the position-based PID controller for the Flapper Drone, we collected offline data by first generating position-based reference trajectories and executing a position-based PID controller to track them.
The position-based reference trajectories are as follows:
\begin{equation}\label{eqn:flapper_reference_trajectory}
    \begin{aligned}
    p_x(t) & \coloneq p_x(0) + (p_x(T_{\max}) - p_x(0)) \frac{t}{T_{\max}} + \varsigma_x \sin(2 \pi \varrho_x t), \\
    p_y(t) & \coloneq p_y(0) + (p_y(T_{\max}) - p_y(0)) \frac{t}{T_{\max}} + \varsigma_y \sin(2 \pi \varrho_y t), \\
    p_z(t) & \coloneq p_z(0) + (p_z(T_{\max}) - p_z(0)) \frac{t}{T_{\max}},
    \end{aligned}
\end{equation}
where $p_x(0), p_y(0), p_z(0)$ and $p_x(T_{\max}), p_y(T_{\max}), p_z(T_{\max})$ represent the initial and final 3D positions, $T_{\max}$ is the maximum flight time, and $\varsigma$ and $\varrho$ denote the amplitude and frequency of the sinusoidal function, respectively. 
Note that we apply the sinusoidal perturbation to only planar positions $p_x$ and $p_y$.
The initial planar positions were sampled uniformly as $p_x, p_y \sim \text{Unif}([-0.3, 0.3])$, while the vertical position was fixed at the robot's nominal height.
Target positions were generated by applying a displacement vector $\Delta p$ to the initial positions, such that $p(T_{\max}) = p(0) + \Delta p$. 
The displacement components were sampled as $\Delta p_x, \Delta p_y \sim \text{Unif}([-0.3, 0.3])$ and $\Delta p_z \sim \text{Unif}([1.0, 1.5])$.
Additionally, we selected small amplitudes $\varsigma_x, \varsigma_y\sim \mathrm{Unif}([0.05, 1])$ and low frequencies $\varrho\sim \mathrm{Unif}([0.01, 0.05])$ to ensure the reference trajectories remain trackable despite high dynamic uncertainties of the robot, and we fixed the yaw angle to zero (i.e., $\psi = 0$).

Following the setup in \Cref{subsubsec:turtlebot_exp_details}, we computed the robot dynamics and collected 26 reference trajectories, 20 of which were used for training, five for validation, and one for testing (evaluation).

\begin{table}[t]
    \centering
    \caption{Identified empirical coefficients for the Flapper Drone dynamics.}
    \label{tab:flapper_empirical_coefficients}
    \begin{tabular}{@{}cc@{\quad}cc@{\quad}cc@{}}
        \toprule
        \textbf{Symbol} & \textbf{Value} & \textbf{Symbol} & \textbf{Value} & \textbf{Symbol} & \textbf{Value} \\
        \midrule
        $c_1$ & 1.0    & $c_5$ & 1.6567 & $c_8$    & 1.0 \\
        $c_2$ & 1.0    & $c_6$ & 6.5546 & $c_9$    & 1.0 \\
        $c_3$ & 1.0    & $b_6$ & 3.1939 & $c_{10}$ & 1.0 \\
        $c_4$ & 1.6321 & $c_7$ & 1.0    &          &     \\
        \bottomrule
    \end{tabular}
\end{table}
\subsubsection{System identification}\label{subsubsec:flapper_system_id}
We modeled the dynamics of the Flapper Drone using the 10D quadrotor model described in \Cref{subsec:simulated_environments} with empirical coefficients as follows:

\begin{equation}\label{eqn:flapper_nominal_dynamics}
    \begin{aligned}
        & f(x) = \qquad\qquad\qquad\qquad B(x) = \\
            &\begin{bmatrix}
                c_1 v_x \\
                c_2 v_y \\
                c_3 v_z \\ 
                c_4  \tilde f \sin(\theta) \\
                -c_5 \tilde f \cos(\theta) \sin(\phi) \\
                c_6 \tilde f \cos(\theta) \cos(\phi) - b_6 \\
                0 \\
                0 \\
                0 \\
                0
                \end{bmatrix},
            \begin{bmatrix}
                0 & 0 & 0 & 0 \\
                0 & 0 & 0 & 0 \\
                0 & 0 & 0 & 0 \\
                0 & 0 & 0 & 0 \\
                0 & 0 & 0 & 0 \\
                0 & 0 & 0 & 0 \\
                c_7 & 0 & 0 & 0 \\
                0 & c_8 & 0 & 0 \\
                0 & 0 & c_9 & 0 \\
                0 & 0 & 0 & c_{10}
                \end{bmatrix}.
    \end{aligned}
\end{equation}

We identified the empirical coefficients using linear least-squares regression. For indices $i \in \{1, 2, 3, 4, 5, 7, 8, 9, 10\}$ (except $6$-th index), we use the standard formulation as in \Cref{eqn:least-squares}.
However, for the $6$-th index where there exists gravitational effect, we include a bias term $b_6$ and solve affine least-squares regression:
\begin{equation}\label{eqn:least-squares_affine}
    c_6, b_6 = \argmin_{c, b \in \mathbb{R}} \sum_{j=1}^{20} \sum_{t=1}^{L_{\max}} \left( \dot{x}_{6, t}^{(j)} - (c \cdot \dot{\hat{x}}_{6, t}^{(j)} + b) \right)^2,
\end{equation}
where $j$ is the trajectory index and $t$ is the time step.
Here, $\dot{x}_{i, t}^{(j)}$ denotes the observed dynamics from the motion capture system (for position and orientation) or computed via finite differences (for the rest of the state indices).
$\dot{\hat{x}}_{i, t}^{(j)}$ denotes the corresponding nominal dynamics term (i.e., the $i$-th row of \Cref{eqn:flapper_nominal_dynamics} excluding the empirical coefficients).
We show the identified empirical coefficients in \Cref{tab:flapper_empirical_coefficients} and visualizations of the least-squares fits in Appendix \ref{app:results} (\Cref{fig:lstsq}b).
These empirical coefficients capture the unique dynamical properties of the Flapper Drone, where the acceleration terms are weighted by coefficients to represent the relationship between the flapping wing motor output and the resulting thrust generation.

Note that the empirical coefficient $b_6$ should correspond to gravitational acceleration, yet it was identified as $3.1939$. 
This discrepancy arises because the Flapper Drone operates on a non-physical hardware thrust scale (10,001–60,000) and uses a normalized thrust in its state.
However, the coefficient is functionally correct as it calibrates the model's hover equilibrium (i.e., $\dot v_z = 0$) to a thrust of $\tilde f = 0.487$, which matches the robot's actual hover thrust.

\subsubsection{Synthesis of path-tracking policies}
We adopted the policy synthesis procedure detailed in \Cref{subsubsec:turtlebot_policy_optimization}.

\subsubsection{Policy deployment}
We executed the policy on a centralized computer at 20 Hz for 30 seconds, while internal sensors and motion capture operated at 100 Hz, thereby avoiding potential sampling instabilities.
We constructed the policy input (state) by using real-time 3D pose data obtained from the motion capture cameras, velocity estimates from the onboard Kalman filter, and a thrust obtained via direct integration ($\tilde f_{t+1} \leftarrow  \tilde f_t + \dot{\tilde{f}} \cdot \Delta t $), where $\dot{\tilde  f}$ is part of the controls output by the policy.
We integrated the thrust because we observed a significant latency in the robot's thrust telemetry.
Finally, we passed the controls output by the policy to the identified system model to generate the target next 3D position $\hat{p}_x, \hat{p}_y, \hat{p}_z$ and yaw angle $ \hat \psi$, which were then passed to the on-board position-based PID controller for execution (see \Cref{sec:cascaded-control-approach}).

\begin{figure*}[t]
    \centering
    \includegraphics[width=0.9\linewidth]{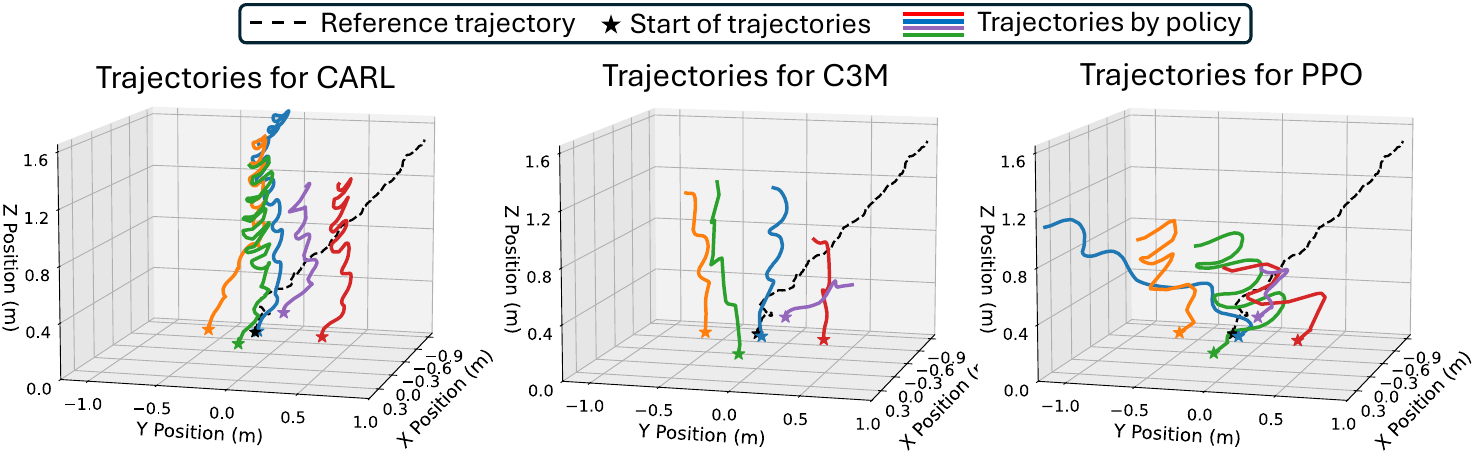}
    \caption{We demonstrated our algorithm (CARL) on the Flapper Drone along with key baselines. The robot was initialized at five distinct starting positions to track the same reference trajectory across all algorithms. 
    }
    \label{fig:flapper_trajectory}
\end{figure*}
\begin{figure*}[t]
    \centering
    \includegraphics[width=0.7\linewidth]{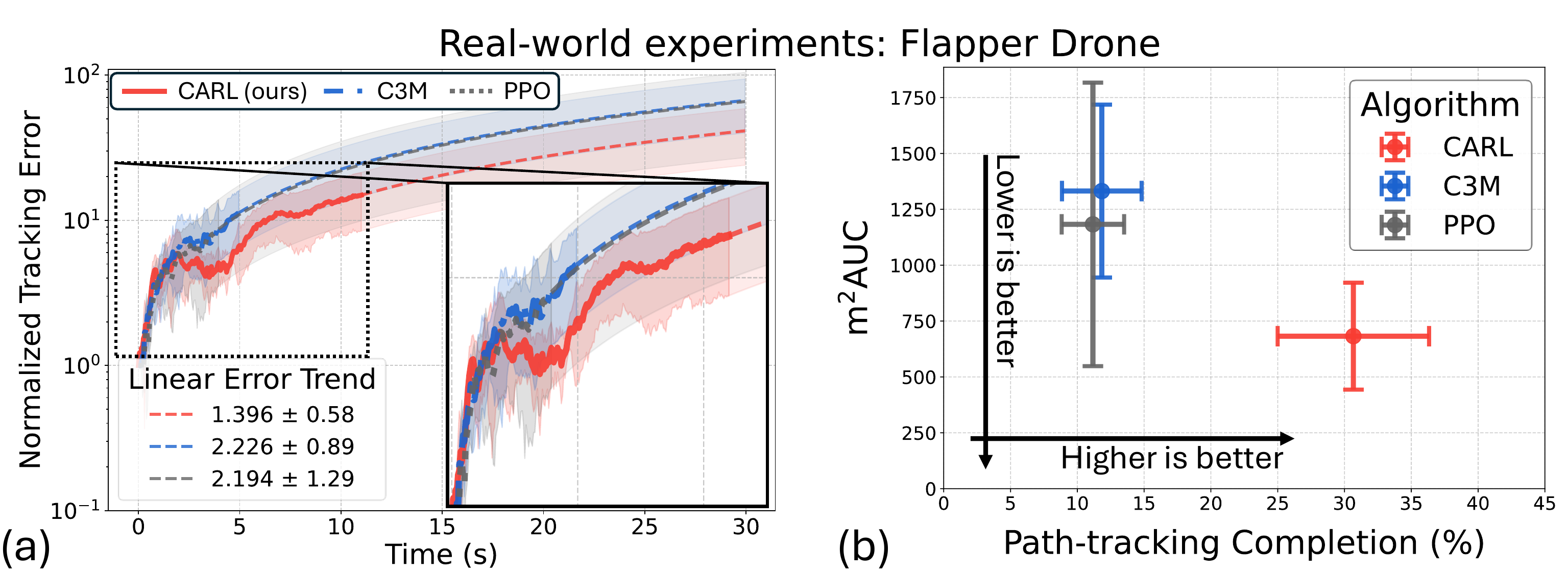}
    \caption{(a) We plot the mean and 95\% confidence intervals of the normalized tracking error over 5 seeds from Flapper experiments. Additionally, (b) we plot the corresponding mean and 95\% confidence intervals of m$^2$AUC and path-tracking completion, where the bottom-right corner denotes the most desirable performance (high completion and low error).
    The higher m$^2$AUC observed, compared to that in \Cref{table:mauc_report}, is attributed to the longer horizon length and increased uncertainty from the sim-to-real transfer.}
    \label{fig:flapper_error}
\end{figure*}

\subsubsection{Evaluation metric}\label{subsubsec:flapper_evaluation_metric}
For these experiments, we slightly revise the mAUC evaluation metric used in \Cref{sec:ground-evaluation-metric} to penalize early termination caused by safety violations, where the robot flew out of a pre-defined safety envelope (which triggered an immediate abort to prevent hardware damage). 
Since early termination represents a critical control failure, we impose an additional penalty for early termination by scaling mAUC by the ratio $\frac{L_{\max}}{L}$. 
The resulting metric, denoted as $\text{m}^2\text{AUC}$, is defined as follows:
\begin{equation}
\text{m}^2\text{AUC} \coloneq \frac{L_{\max}}{L} \cdot \underbrace{\frac{L_{\max}}{L} \sum_{k=0}^{L} \|\mathrm e(t_k)\|_2}_{\text{mAUC}}.
\end{equation}

Additionally, we provide a linear error trend ($\tau^\star$) and path-tracking completion ($\iota$) for a comprehensive assessment under early termination, where $\tau$ and $\iota$ are defined as follows:
\begin{align}
    &\qquad  \tau^\star \coloneq \argmin_{\tau} \sum_{k=0}^{L} \left( \|\mathrm{e}(t_k)\|_2 - (\tau t_k + 1) \right)^2, \\
    & \qquad  \iota \coloneq \frac{L}{L_{\max}} \times 100\%.
\end{align}
The bias term in finding $\tau^\star$ is constrained to $1$ to strictly measure the linear error trend, reflecting the fact that the initial normalized tracking error is always $1$.

\subsection{Flapper Drone Results}
\Cref{fig:flapper_trajectory} shows the reference trajectory (i.e., test trajectory) and executed trajectories from deploying trained policies on the Flapper Drone with five different initial states for CARL, C3M, and PPO.
\Cref{fig:flapper_error}a shows corresponding normalized tracking errors with linear error trend and \Cref{fig:flapper_error}b plots the m$^2$AUC and path-tracking completion achieved by each algorithm.
While the sim-to-real gap introduces large tracking errors across all algorithms, CARL demonstrates superior performance and robustness relative to baselines. 
For example, \Cref{fig:flapper_error} shows that CARL outperforms the baselines across all evaluated metrics: m$^2$AUC, linear error trend, and tracking completion.
The short path-tracking completions for C3M and PPO are due to erroneous control inputs that drive the robot outside the safety region.
Conversely, CARL produces controls that better regulate the Flapper Drone to the reference trajectory, mitigating the risk of early termination and achieving the best path-tracking performance.

\section{Real-World Experiment: Jethexa}\label{sec:experiment_legged_robot}
To scale beyond low-dimensional mobile robots such as TurtleBot3 Burger and Flapper Drone, we evaluated CARL on the \href{https://www.hiwonder.com/products/jethexa}{Jethexa} hexapod---which features complex kinematic coupling between state elements---depicted in \Cref{fig:experiment_facility}b, within the same facility shown in \Cref{fig:experiment_facility}a.
We followed the same experimental pipeline shown in \Cref{fig:cac_high_level}.

\subsection{Jethexa Experimental Details}

\subsubsection{Offline data collection}\label{subsubsec:jethexa_offline_data_collection}
The Jethexa platform provides an inverse kinematics solver that translates given velocity commands into a desired locomotion. 
We used this approach to generate reference trajectories by sampling a forward linear velocity $v_x \sim \text{Unif}(0.02, 0.05)$, while holding the lateral velocity $v_y$ zero. 
Additionally, we included an angular velocity using a sinusoidal function as $\omega = \varsigma \sin(2\pi \varrho )$ where we sampled $\varsigma\sim \text{Unif}(0.02, 0.05)$ and a frequency of $\varrho \coloneq 2 \pi / T_{\max}$, ensuring it completes exactly one full cycle while generating a diverse set of reference trajectories.
We use the motion capture system in our facility (\Cref{fig:experiment_facility}a) to record the robot's base position and attitude, while joint angles are read directly from the robot.
Similarly, we use the second-order finite difference method to compute the dynamics $\dot x$.

Following the setup in \Cref{subsubsec:turtlebot_exp_details}, we computed the robot dynamics and collected 26 reference trajectories, 20 of which were used for training, five for validation, and one for testing (evaluation).

\begin{figure*}[t]
    \centering
    \includegraphics[width=0.95\linewidth]{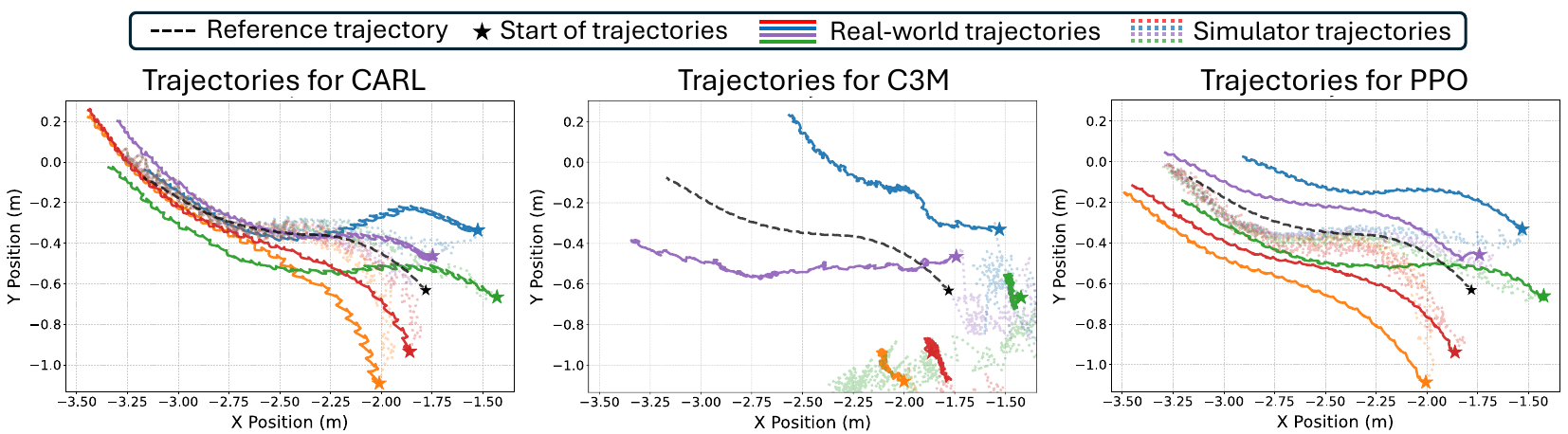}
    \caption{We demonstrated our algorithm (CARL) on the Jethexa along with key baselines. The robot was initialized at five distinct starting positions to track the same reference trajectory across all algorithms. 
    }
    \label{fig:jethexa_traj}
\end{figure*}

\begin{figure}[t]
    \centering
    \includegraphics[width=0.95\linewidth]{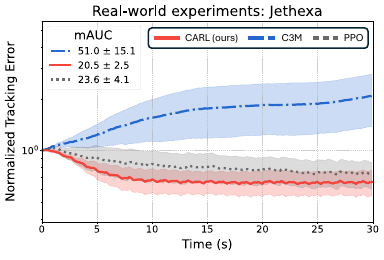}
    \caption{Mean and 95\% confidence intervals for the mAUC from simulation experiments. The higher mAUC observed, compared to that in \Cref{table:mauc_report}, is attributed to the longer horizon length and increased uncertainty from the sim-to-real transfer.}
    \label{fig:jethexa_error}
\end{figure}

\begin{figure}[t]
    \centering
    \includegraphics[width=0.9\linewidth]{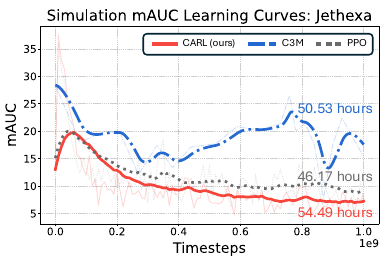}
    \caption{Learning curves of the mAUC for each algorithm during Jethexa simulation experiments. The total wall-clock time for training is denoted. }
    \label{fig:jethexa_learning_curve}
\end{figure}

\subsubsection{System identification}\label{subsubsec:jethexa_system_id}
We modeled the dynamics of the Jethexa as follows:
\begin{equation}\label{eqn:jethexa_nominal_dynamics}
    \begin{aligned}
        f(x) = \mathbf{0}_{23 \times 1}, \;
            B(x) = \begin{bmatrix}
                \mc{J}_{\varpi}(x) \\
                I_{18 \times 18}
                \end{bmatrix},
    \end{aligned}
\end{equation}
where the state is defined as $x \coloneq (p_x, p_y, \phi, \theta, \psi, \mathbf{q}_{18})$, with $p_x$ and $p_y$ denoting the 2D position, $\phi, \theta$, and $\psi$ are the attitude, and $\mathbf{q}_{18}$ denotes the 18 joint angles.
Specifically, the model consists of six legs, each featuring three joints (the coxa, femur, and tibia), resulting in 18 joint angles.
The control input is defined as $u \coloneq \dot{\mathbf{q}}_{18}$, representing angular velocities of each joint. 
The base dynamics $\mc{J}_{\varpi}(x)$ is produced for each state $x$ by a feed-forward neural network with parameters $\varpi$ trained on training data (see \Cref{subsubsec:jethexa_offline_data_collection}) as described in \Cref{sec:pretraining-dynamics-model}.
$\varpi$ was represented by hidden layers of size (64, 64) and Tanh activation functions. 
We visualize the prediction results of $\mc{J}_{\varpi}(x)$ on the dynamics of the test trajectories and the corresponding mean-squared error in Appendix \ref{app:results} (\Cref{fig:lstsq}c).

\subsubsection{Synthesis of path-tracking policies}
We adopted the policy synthesis procedure detailed in \Cref{subsubsec:turtlebot_policy_optimization}.

\subsubsection{Policy deployment}

We executed the policy on a centralized computer at 20 Hz for 30 seconds, while internal sensors operated at 50 Hz and motion capture at 100 Hz.
We constructed the policy input (state) using real-time data obtained from motion capture cameras and joint angles from the onboard sensor.
We integrated the control outputs (i.e., joint angle velocities) by the policy as $\mathbf{q}_{18} \leftarrow \mathbf{q}_{18} + \dot{\mathbf{q}}_{18} \Delta t $ and clipped to a safe operational range defined by 10\% expansion of the joint angle range found in the training data.
Finally, the clipped joint angle target is then commanded to the onboard PID controllers.

\subsubsection{Evaluation metric}
We adopted the same evaluation procedure established for the TurtleBot3 Burger in \Cref{sec:ground-evaluation-metric}.

\subsection{Jethexa Results}

\Cref{fig:jethexa_traj} shows the reference trajectory (i.e., test trajectory) and executed trajectories from deploying trained policies on the Jethexa with five different initial states for CARL, C3M, and PPO.
We illustrate the sim-to-real gap by including open-loop predicted trajectories (shown as dotted lines) that represent the policy's expected path from each initial state under identified dynamics in \Cref{eqn:jethexa_nominal_dynamics}.
\Cref{fig:jethexa_error} shows the corresponding normalized tracking errors for these trajectories, with the mean and 95\% confidence intervals.
The mAUC learning curves and corresponding total wall-clock time for training of each algorithm are also depicted in \Cref{fig:jethexa_learning_curve}.

We observe that CARL outperforms both C3M and PPO in real-world deployment and simulation experiment as depicted in \Cref{fig:jethexa_error} and \Cref{fig:jethexa_learning_curve}, respectively. 
As seen in \Cref{fig:jethexa_traj}, CARL adopts a slightly more aggressive control strategy to minimize cumulative tracking error in a contracting manner to effectively track the reference trajectory. 
In contrast, C3M fails to maintain stable tracking---with several trials diverging significantly---while PPO, despite its smoother control profile, exhibits noticeable path-tracking error compared to CARL.
Finally, we see that CARL shows an increase in training wall-clock time of only 8\% relative to C3M and 18\% relative to PPO.

We note that both CARL and the baselines achieve better real-world tracking performance on the Jethexa compared to the TurtleBot3 Burger and Flapper Drone. 
Two potential reasons for this improvement are that the Jethexa is more robust to uncertainties, like friction and latency dynamics, which reduces the estimation error due to data scarcity in system identification, and that we used a neural network to model its dynamics, rather than least-squares.

\section{Conclusions} \label{sec:conclusion}
We propose a contraction-aware RL algorithm, CARL, for robust path-tracking in high-dimensional nonlinear systems under approximate dynamics. Our key idea is to simultaneously learn a CMG and a policy, where the policy optimizes a CCM-conditioned reward function defined by CCMs generated by the CMG. 
Our resulting policy achieves stability and long-term optimality, as well as robustness to dynamics approximation errors.
While our results use PPO for RL policy optimization, in principle, CARL can be implemented with any RL algorithm.
We empirically demonstrate the benefits of CARL relative to baselines through extensive simulated and real-world experiments.
We also theoretically analyze the stability of CARL and its robustness under approximate dynamics.

One limitation of CARL is its reliance on a simulator for RL training.
However, recent progress in robotics has resulted in many high-fidelity simulators, such as \href{https://developer.nvidia.com/isaac/sim}{NVIDIA Isaac Sim} and \href{https://genesis-embodied-ai.github.io/}{Genesis}, which provide fast, realistic, and easily-integrable environments for training a policy.

We also outline several promising directions for future work. 
First is extending CARL to the offline RL setting, which would address our reliance on a simulator. 
Second is considering robust RL formulations---such as domain randomization or mini-max worst-case optimization---to further improve the sim-to-real gaps shown in our robot experiments and compare with additional baselines (e.g., well-tuned classical controllers).
These additions should better clarify the practical significance of our work.
Other directions include jointly training all models (dynamics, CMG, and policy).

\section*{Acknowledgments}
Listed in alphabetical order, the authors would like to thank Rihan D’silva, Inyoung Kim, and Hamid Osooli for their insightful discussions and valuable assistance with the real-world robot experiments.

This paper will appear in the IEEE Transactions on Robotics. © 2026 IEEE. Personal use of this material is permitted. Permission from IEEE must be obtained for all other uses, in any current or future media, including reprinting/republishing this material for advertising or promotional purposes, creating new collective works, for resale or redistribution to servers or lists, or reuse of any copyrighted component of this work in other works.

\appendices

\section{Omitted Proofs}\label{app:proofs}

\begin{proofof}{\Cref{lemma:equivalence_of_optimizations}}\label{proof:lem1}
    Given $\tilde{C}(s) = 1 - R(s)$, the objective based on the reward value function of any policy $\pi$ is given as follows:
    \begin{align}
        J_{R}(\pi)
        &= \mathbb{E}_{s_0} \left[ \mathbb{E}_{\pi} \left[ \sum_{t=0}^\infty \gamma^t (1 - \tilde{C}(s_t)) \mid s_0 \right]\right] \\
        &= \frac{1}{1-\gamma} - J_{\tilde{C}}(\pi),
    \end{align}
    where we used geometric series as $\gamma < 1$ ensures the series converge.
    Given that the term $(1-\gamma)^{-1}$ is constant, and assuming consistent tie-breaking rules, the optimization problem simplifies to
    \begin{align}
        \pi^\star 
        &= \argmax_\pi \left( \frac{1}{1 - \gamma } - J_{\tilde{C}}(\pi) \right) \\
        &= \argmax_\pi \left( - J_{\tilde{C}}(\pi) \right)= \argmin_\pi J_{\tilde{C}}(\pi).
    \end{align}
\end{proofof}

\begin{proofof}{\Cref{lemma:contraction_in_C_tilde}}\label{proof:lem2}
    Let $\tau_k \coloneq 2\lambda k \Delta t$ for notational simplicity. By the strict monotonicity of $\tilde{C}$ and the original contraction bound $C(s_{t+k}) \leq C(s_t) e^{-\tau_k}$, we have:
    \begin{align}
        \tilde{C}(s_{t+k}) &\leq \frac{C(s_t) e^{-\tau_k}}{1 + C(s_t) e^{-\tau_k}} \\
        &= \tilde{C}(s_t) \left( \frac{1 + C(s_t)}{1 + C(s_t)e^{-\tau_k}} \right) e^{-\tau_k} \\
        &= \tilde{C}(s_t) \exp\left( -\tau_k + \ln\frac{1 + C(s_t)}{1 + C(s_t)e^{-\tau_k}} \right). \label{eqn:tilde_C_bound}
    \end{align}
    Defining the $k$-th step contraction rate $\hat{\lambda}_k$ such that the exponent in \Cref{eqn:tilde_C_bound} equals $-2\hat{\lambda}_k k \Delta t$, we obtain:
    \begin{equation}
        \hat{\lambda}_k = \lambda \left( 1 - \frac{1}{\tau_k} \ln{\frac{1 + C(s_t)}{1 + C(s_t)e^{-\tau_k} } } \right).
    \end{equation}
    Because $e^{-\tau_k} < 1$, it trivially holds that 
    \begin{equation}
        0 < \ln \frac{1 + C(s_t)}{1 + C(s_t)e^{-\tau_k}} < {\tau_k},
    \end{equation}
    ensuring $\hat{\lambda}_k \in (0, \lambda)$. 
    Let $\tilde{\lambda} \coloneq \inf_{k \in\mb{N}} \hat{\lambda}_k > 0$.
    Then, there exists a rate $\tilde{\lambda} \in (0, \lambda)$ such that $\pi_c$ contracts with respect to $\tilde{C}$ as follows:
    \begin{equation}
        \tilde{C}(s_{t+k}) \leq \tilde{C}(s_t) e^{-2\tilde{\lambda} k \Delta t}, \; \forall k \in \mb{N}.
    \end{equation}
\end{proofof}

We establish the following lemma that we extensively use to prove the main theorems.
\begin{lemma}\label{lemma:integration_of_contraction_inequality}
\emph{(Uniformly Bounded Cumulative Cost of an optimal Policy).} 
Suppose there exists at least one policy $\pi_c$ that contracts with respect to $C$ with rate $\lambda >0$ as in \Cref{assumption:1}.
Let the cost function and corresponding cumulative cost be defined as in \Cref{eqn:new_cost_function} and \Cref{eqn:new_cost_value_function}, respectively.
Then, the cost value function of a deterministic optimal policy $\pi^\star$ satisfying \Cref{eqn:discounted_optimal_policy} is uniformly bounded as follows:

\begin{equation}
    V_{\tilde{C}}^{\pi^\star}(s_t) \leq V_{\tilde{C}}^{\pi_c}(s_t) =\sum_{k=0}^{\infty} \gamma^k \tilde{C}(s_{t+k}) \leq \frac{ \tilde{C}(s_t) }{1 - \gamma e^{-2 \tilde{\lambda} \Delta t}},
\end{equation}
where $\tilde{\lambda}$ is the contraction rate induced by $\pi_c$ with respect to the cost function $\tilde{C}$ as in \Cref{lemma:contraction_in_C_tilde}.

\end{lemma}

\begin{proofof}{\Cref{lemma:integration_of_contraction_inequality}}\label{proof:lem3}
    By \Cref{lemma:contraction_in_C_tilde}, there exists a rate $\tilde{\lambda} \in (0, \lambda)$ such that $\pi_c$ contracts with respect to $\tilde{C}$.
    Thus, using the optimality of $\pi^\star$ and the fact that the cost value function of $\pi_c$ is uniformly bounded using geometric series, we get the following relationship:
    \begin{align}
        V_{\tilde{C}}^{\pi^\star}(s_t) \leq V_{\tilde{C}}^{\pi_c}(s_t) = \sum_{k=0}^{\infty} \gamma^k \tilde{C}(s_{t+k}) \leq \frac{\tilde{C}(s_t)}{1 - \gamma e^{-2 \tilde{\lambda} \Delta t}}.
    \end{align}

\end{proofof}

\begin{proofof}{\Cref{thm:asymptotic_stability}}
    We start from the Bellman equation of the optimal policy $\pi^\star$. 
    Given  \Cref{lemma:equivalence_of_optimizations}, we write the equation in terms of cost using $V^{\pi^\star}_{\tilde C}(s_t)$:
    \begin{equation}\label{eqn:Bellman_optimal_cost_eqn}
        V^{\pi^\star}_{\tilde C}(s_t) = \tilde C(s_t) + \gamma V^{\pi^\star}_{\tilde C}(s_{t+1}),
    \end{equation}
    where the expectation operator is omitted from the RHS, as we focus on deterministic policy and dynamics.
    Rearranging \Cref{eqn:Bellman_optimal_cost_eqn} isolates the difference in the cost value function:
    \begin{equation}\label{eqn:rearranged_Bellman_optimal_cost_eqn}
        V^{\pi^\star}_{\tilde C}(s_{t+1}) - V^{\pi^\star}_{\tilde C}(s_t) = \frac{1}{\gamma} \left( (1 - \gamma) V^{\pi^\star}_{\tilde C}(s_t) - \tilde C(s_t) \right).
    \end{equation}
    For the value function to strictly decrease, the term inside the parentheses on the RHS must be negative, which yields the following condition:
    \begin{equation}\label{eqn:sufficient_condition_of_negative_definitness}
          V^{\pi^\star}_{\tilde C}(s_t) < \frac{\tilde{C}(s_t)}{1 - \gamma}.
    \end{equation}
    To prove the existence of $\gamma \in (0, 1)$ that satisfies \Cref{eqn:sufficient_condition_of_negative_definitness}, we bound the LHS of \Cref{eqn:sufficient_condition_of_negative_definitness} by a finite measure while we let the RHS approach infinity as $\gamma \rightarrow 1$.

    Using \Cref{lemma:integration_of_contraction_inequality}, we substitute the upper bound of the cost value function of the contracting policy $\pi_c$ to the LHS of \Cref{eqn:sufficient_condition_of_negative_definitness} as follows:
    \begin{equation}\label{eqn:semi_sufficient_condition_of_gamma}
        \frac{\tilde{C}(s_t)}{1 - \gamma e^{-2 \tilde{\lambda} \Delta t}} < \frac{\tilde C(s_t)}{1 - \gamma} \implies e^{-2 \tilde{\lambda}\Delta t} < 1,
    \end{equation}
    which proves the cost value function of $\pi^\star$ is a strictly decreasing sequence  (i.e., $\{V^{\pi^\star}_{\tilde C}(s_t)\}_{t=0}^\infty$) for $\gamma \in (0, 1)$.

    Given that the value function is bounded below by 0, it converges to some limit $L \geq 0$ by the monotone convergence theorem \cite{royden1988real}. 
    Taking $t \to \infty$ in the Bellman equation from \Cref{eqn:Bellman_optimal_cost_eqn} gives
    \begin{equation}\label{eqn:Bellman_limit}
        \lim_{t\to\infty} \tilde{C}(s_t) = (1-\gamma)L.
    \end{equation}
    Again, applying \Cref{lemma:integration_of_contraction_inequality} to \Cref{eqn:Bellman_limit} yields 
    \begin{equation}
        L \leq (1-\gamma)L/(1 - \gamma e^{-2\tilde{\lambda}\Delta t}).
    \end{equation}
    Suppose $L > 0$. Dividing both sides by $L$ and simplifying yields $e^{-2\tilde{\lambda}\Delta t} \geq 1$, which contradicts the fact that $\tilde{\lambda} > 0$ and $\Delta t > 0$.
    Hence $L = 0$, implying incremental asymptotic stability:
    \begin{equation}
        \lim_{k\to\infty} \| \delta x(t_k)\|_M^2 = 0, \quad \forall \delta x(t_0)\in\mb{R}^n.
    \end{equation}
\end{proofof}

\begin{proofof}{\Cref{thm:exponential_stability}}
    By factoring the rearranged Bellman equation (\Cref{eqn:Bellman_optimal_cost_eqn}) with respect to $V^{\pi^\star}_{\tilde C}(s_t)$, we have
    \begin{equation}\label{eqn:eta_definition_proof}
        V^{\pi^\star}_{\tilde C}(s_{t+1}) = V^{\pi^\star}_{\tilde C}(s_t) \left[ \frac{1}{\gamma} \left( 1 - \frac{\tilde C(s_t)}{V^{\pi^\star}_{\tilde C}(s_t)} \right) \right]  =  V^{\pi^\star}_{\tilde C}(s_t) \rho_t,
    \end{equation}
    where $\rho_t$ is a contraction factor (i.e., $\rho_t < 1, \; \forall t$) by using $\gamma \in (0, 1)$ (see the proof of \Cref{thm:asymptotic_stability}).
    
    Given that the value function is always lower bounded by its instantaneous cost and upper bounded as shown in \Cref{lemma:integration_of_contraction_inequality}, we establish the following inequality based on \Cref{eqn:eta_definition_proof}:
    \begin{equation}\label{eqn:recursive_V_bound}
        \tilde{C}(s_{t+1}) \leq V^{\pi^\star}_{\tilde C}(s_{t+1}) = V^{\pi^\star}_{\tilde C}(s_{t})  \rho_{t} \leq  \tilde{C}(s_{t}) \frac{\rho_{t}}{1 - \gamma e^{-2\tilde{\lambda}\Delta t}}.
    \end{equation}

    Let us define $\hat{\rho}_t \coloneq \frac{\rho_{t}}{1 - \gamma e^{-2\tilde{\lambda}\Delta t}}$ and $\overline{\rho} \coloneq \sup_{t}\hat{\rho}_t$.
    If $\overline{\rho} < 1$, then a deterministic optimal policy $\pi^\star$ exhibits incremental exponential stability as follows:
    \begin{equation}\label{eqn:saturated_cost_contraction}
        \tilde{C}(s_{t+1}) \leq \tilde{C}(s_{t}) \hat{\rho}_{t} \leq \tilde{C}(s_{0}) \left(\prod_{i=0}^{t} \hat{\rho}_{i} \right) \leq  \tilde{C}(s_{0}) \overline{\rho}^{t+1}.
    \end{equation}
    Since $\tilde{C}(s_t)$ is strictly monotonic, the contraction of $\tilde{C}(s_t)$ in \Cref{eqn:saturated_cost_contraction} implies contraction with respect to $C(s_t)$ for some contraction factor.
    Hence, the following holds:
    \begin{align}
        & \frac{{C}(s_{t+1})}{1 + {C}(s_{t+1})} \leq   \frac{{C}(s_{t})}{1 + {C}(s_{t})}\overline{\rho},  \quad \forall t \\
        & \implies  C(s_{t+1}) \leq  \frac{{C}(s_{t})(1 + C(s_{t+1}) )}{1 + {C}(s_{t})}  \overline{\rho} \leq C(s_t) \overline{\rho}  , \; \forall t, \label{eqn:cost_contraction}
    \end{align}
    which directly implies incremental exponential stability as follows:
    \begin{align}
        \| \delta x(t_k) \|_M^2 \leq \| \delta x(t_0) \|_M^2 \overline{\rho}^{k}, \; \forall k \in \mb{N}.
    \end{align}

    Next, we show when $\overline{\rho} < 1$ is true using the following condition:
    \begin{align}
        \hat{\rho}_t = \frac{1}{\gamma(1 - \gamma e^{-2\tilde{\lambda } \Delta t})} \left( 1 - \frac{\tilde C(s_t)}{V^{\pi^\star}_{\tilde C}(s_t)} \right) < 1,\quad \forall t,
    \end{align}
    which simplifies to showing
    \begin{equation}\label{eqn:condition_for_exp_stability}
         1 - \frac{\tilde C(s_t)}{V^{\pi^\star}_{\tilde C}(s_t)} < \gamma(1 - \gamma e^{-2\tilde{\lambda }\Delta t}), \quad 
         \forall t.
    \end{equation}
    Applying \Cref{lemma:integration_of_contraction_inequality} to \Cref{eqn:condition_for_exp_stability} simplifies the condition for incremental exponential stability as follows:
    \begin{equation}\label{eqn:condition_for_exp_stability_of_RL_policy}
        \gamma  e^{-2\tilde{\lambda } \Delta t} < \gamma (1 - \gamma  e^{-2\tilde{\lambda } \Delta t}) \implies  e^{-2\tilde{\lambda } \Delta t} < \frac{1}{1 + \gamma}.
    \end{equation}

    Rewriting \Cref{eqn:condition_for_exp_stability_of_RL_policy} in terms of $\gamma$, we define the threshold on the discount factor for incremental exponential stability as follows:
    \begin{equation}
        \overline{\gamma} \coloneq e^{2\tilde{\lambda}\Delta t} - 1 > 0.
    \end{equation}

    Hence, for a discount factor $\gamma < \min(\overline{\gamma}, 1)$, there exists a contraction factor $\overline{\rho} \in (0, 1)$ such that a deterministic optimal policy $\pi^\star$ satisfying \Cref{eqn:discounted_optimal_policy} renders the closed-loop system incrementally exponentially stable.

\end{proofof}

\begin{proofof}{\Cref{thm_robustness}}
As shown in~\cite{tingwei_conformal} for Lyapunov functions, the contracting policy $\pi_c$ yields
\begin{align}
    \label{eq__pointwise_cp_bound}
    \mb{P}\left[\sqrt{C(s_k)} \leq \sqrt{C(s_0)} e^{-\lambda k \Delta t}+\frac{\sqrt{\overline{m}}\mf{Q}_{1-\varepsilon}}{\lambda}\right] \geq 1-\varepsilon.
\end{align}
Since $\pi_c$ is deterministic and we have $\sqrt{\sum_{k=0}^{\kappa-1} C(s_k)} \leq \sum_{k=0}^{\kappa-1} \sqrt{C(s_k)}$ for any nonnegative sequence $\{C(s_k)\}$, summing the event of~\Cref{eq__pointwise_cp_bound} over $k$ and applying geometric series gives
\begin{align}
    \sqrt{\mb{E}_{\pi_c} \left[ \sum_{k=0}^{\kappa-1} C(s_{k}) \right]} \leq \underbrace{\frac{(1-e^{-\lambda\kappa\Delta t})\sqrt{C(s_0)}}{1-e^{-\lambda\Delta t}}+\frac{\sqrt{\overline{m}}\mf{Q}_{1-\varepsilon}\kappa}{\lambda}}_{\coloneq\textbf{UB} }.
\end{align}
Thus, assuming $\pi^\star$ achieves lower cumulative tracking error than $\pi_c$, taking limit $\kappa \rightarrow \infty$ gives the following bound for the deterministic optimal policy $\pi^\star$:
\begin{align}
    \mb{P}\left[\lim_{\kappa \rightarrow \infty} \tfrac{\sqrt{\mb{E}_{\pi^\star} \left[ \sum_{k=0}^{\kappa-1} C(s_k) \right]}}{\kappa} \leq \frac{ \textbf{UB}}{\kappa}\right] \geq 1-\varepsilon.
\end{align}
Since $\pi^\star$ is deterministic, this implies~\Cref{eq_robust_bound}.
\end{proofof}

\section{Additional Results}
\label{app:results}

\begin{figure*}[t]
    \centering
    \includegraphics[width=0.98\linewidth]{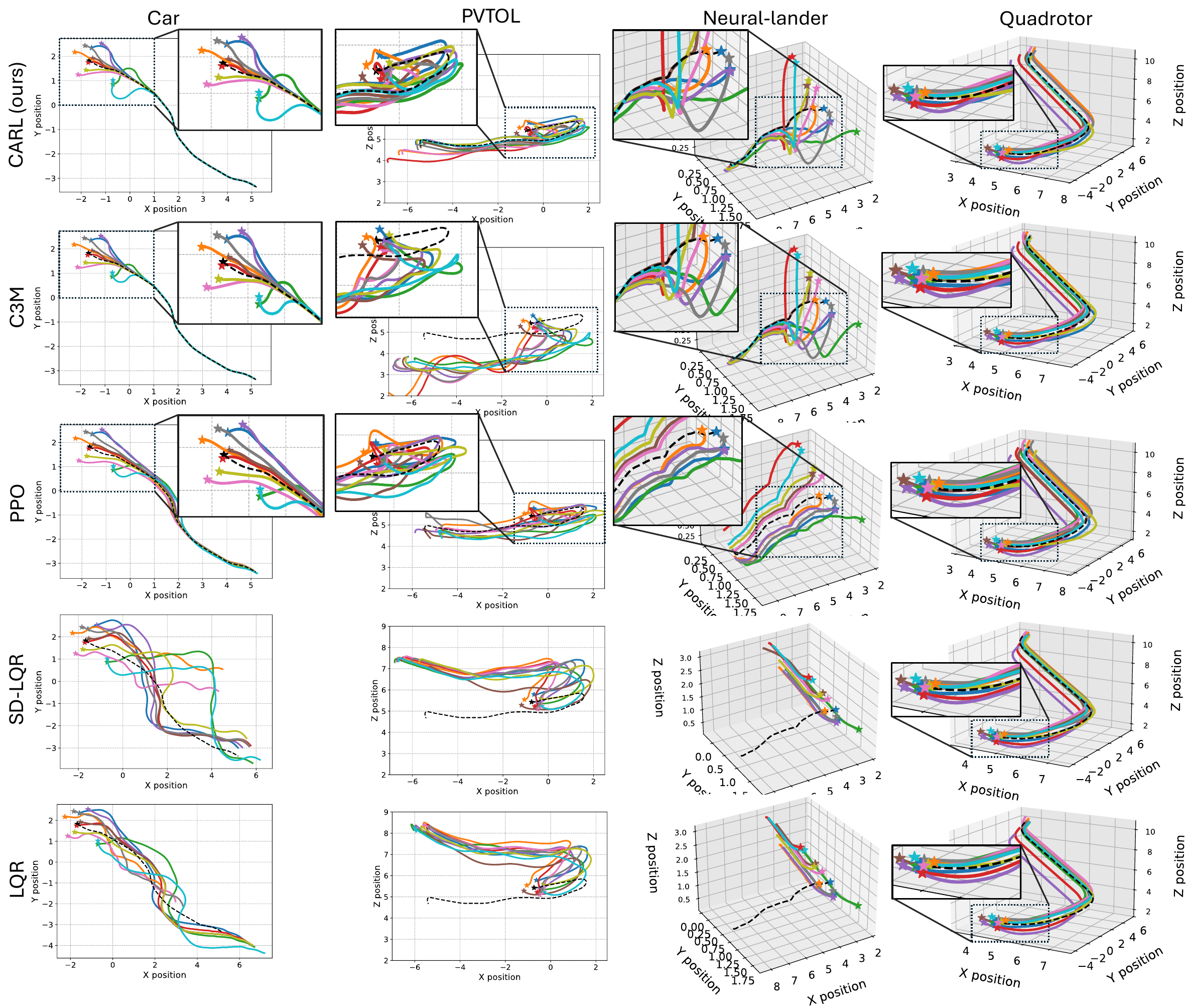}
    \caption{We plot simulated trajectories of final policies for each algorithm, including magnified insets to highlight stabilizing behaviors. 
    We omit these insets for the LQR-based methods in the CAR, PVTOL, and Neural-lander environments, as their divergence is already clearly visible at the standard scale.}
    \label{fig:simulation_trajectory}
\end{figure*}

\begin{figure*}[t]
    \centering
    \includegraphics[width=0.99\linewidth]{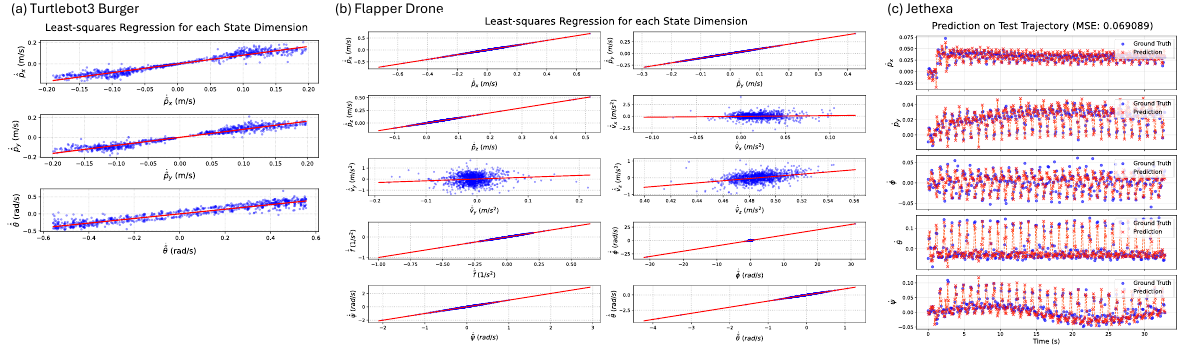}
    \caption{We plot the results of system identification results for the TurtleBot3 Burger, Flapper Drone, and Jethexa. The identified empirical coefficients for the TurtleBot3 Burger and the Flapper Drone are listed in \Cref{tab:hyperparameters}, and the identified dynamics model ($\mc{J}_{\varpi}$) of Jethexa is used in \Cref{eqn:jethexa_nominal_dynamics}.}
    \label{fig:lstsq}
\end{figure*}

We depict path-tracking results of CARL and baseline algorithms in \Cref{fig:simulation_trajectory} for the simulated environments. 
We plot the system identification results for the TurtleBot3 Burger, Flapper Drone, and Jethexa in \Cref{fig:lstsq}.

\bibliographystyle{IEEEtran}
\bibliography{ref}

@article{LOHMILLER1998683,
title = {On Contraction Analysis for Non-linear Systems},
journal = {Automatica},
volume = {34},
number = {6},
pages = {683-696},
year = {1998},
issn = {0005-1098},
doi = {https://doi.org/10.1016/S0005-1098(98)00019-3},
url = {https://www.sciencedirect.com/science/article/pii/S0005109898000193},
author = {WINFRIED Lohmiller and JEAN-JACQUES E. Slotine}
}

@article{tsukamoto2021contraction,
  title={Contraction theory for nonlinear stability analysis and learning-based control: A tutorial overview},
  author={Tsukamoto, Hiroyasu and Chung, Soon-Jo and Slotine, Jean-Jaques E},
  journal={Annual Reviews in Control},
  volume={52},
  pages={135--169},
  year={2021},
  publisher={Elsevier}
}

@article{manchester2017control,
  title={Control contraction metrics: Convex and intrinsic criteria for nonlinear feedback design},
  author={Manchester, Ian R and Slotine, Jean-Jacques E},
  journal={IEEE Transactions on Automatic Control},
  volume={62},
  number={6},
  pages={3046--3053},
  year={2017},
  publisher={IEEE}
}

@book{puterman2014markov,
  title={Markov decision processes: discrete stochastic dynamic programming},
  author={Puterman, Martin L},
  year={2014},
  publisher={John Wiley \& Sons}
}

@inproceedings{sun2021learning,
  title={Learning certified control using contraction metric},
  author={Sun, Dawei and Jha, Susmit and Fan, Chuchu},
  booktitle={conference on Robot Learning},
  pages={1519--1539},
  year={2021},
  organization={PMLR}
}

@inproceedings{song2024stable,
  title={Stable modular control via contraction theory for reinforcement learning},
  author={Song, Bing and Slotine, Jean-Jacques and Pham, Quang-Cuong},
  booktitle={6th Annual Learning for Dynamics \& Control Conference},
  pages={1136--1148},
  year={2024},
  organization={PMLR}
}

@article{tsukamoto2021learning,
  title={Learning-based robust motion planning with guaranteed stability: A contraction theory approach},
  author={Tsukamoto, Hiroyasu and Chung, Soon-Jo},
  journal={IEEE Robotics and Automation Letters},
  volume={6},
  number={4},
  pages={6164--6171},
  year={2021},
  publisher={IEEE}
}

@article{tsukamoto2020neural,
  title={Neural contraction metrics for robust estimation and control: A convex optimization approach},
  author={Tsukamoto, Hiroyasu and Chung, Soon-Jo},
  journal={IEEE Control Systems Letters},
  volume={5},
  number={1},
  pages={211--216},
  year={2020},
  publisher={IEEE}
}

@article{singh2023robust,
  title={Robust feedback motion planning via contraction theory},
  author={Singh, Sumeet and Landry, Benoit and Majumdar, Anirudha and Slotine, Jean-Jacques and Pavone, Marco},
  journal={The International Journal of Robotics Research},
  volume={42},
  number={9},
  pages={655--688},
  year={2023},
  publisher={SAGE Publications Sage UK: London, England}
}

@inproceedings{richards2023learning,
  title={Learning control-oriented dynamical structure from data},
  author={Richards, Spencer M and Slotine, Jean-Jacques and Azizan, Navid and Pavone, Marco},
  booktitle={International Conference on Machine Learning},
  pages={29051--29062},
  year={2023},
  organization={PMLR}
}

@inproceedings{shengyi2022the37implementation,
  author = {Huang, Shengyi and Dossa, Rousslan Fernand Julien and Raffin, Antonin and Kanervisto, Anssi and Wang, Weixun},
  title = {The 37 Implementation Details of Proximal Policy Optimization},
  booktitle = {ICLR Blog Track},
  year = {2022},
  note = {https://iclr-blog-track.github.io/2022/03/25/ppo-implementation-details/},
  url  = {https://iclr-blog-track.github.io/2022/03/25/ppo-implementation-details/}
}

@inproceedings{engstrom2019implementation,
  title={Implementation matters in deep rl: A case study on ppo and trpo},
  author={Engstrom, Logan and Ilyas, Andrew and Santurkar, Shibani and Tsipras, Dimitris and Janoos, Firdaus and Rudolph, Larry and Madry, Aleksander},
  booktitle={International conference on learning representations},
  year={2019}
}

@inproceedings{andrychowicz2021matters,
  title={What matters for on-policy deep actor-critic methods? a large-scale study},
  author={Andrychowicz, Marcin and Raichuk, Anton and Sta{\'n}czyk, Piotr and Orsini, Manu and Girgin, Sertan and Marinier, Rapha{\"e}l and Hussenot, Leonard and Geist, Matthieu and Pietquin, Olivier and Michalski, Marcin and others},
  booktitle={International conference on learning representations},
  year={2021}
}

@article{schulman2017proximal,
  title={Proximal policy optimization algorithms},
  author={Schulman, John and Wolski, Filip and Dhariwal, Prafulla and Radford, Alec and Klimov, Oleg},
  journal={arXiv preprint arXiv:1707.06347},
  year={2017}
}

@article{singh2021learning,
  title={Learning stabilizable nonlinear dynamics with contraction-based regularization},
  author={Singh, Sumeet and Richards, Spencer M and Sindhwani, Vikas and Slotine, Jean-Jacques E and Pavone, Marco},
  journal={The International Journal of Robotics Research},
  volume={40},
  number={10-11},
  pages={1123--1150},
  year={2021},
  publisher={SAGE Publications Sage UK: London, England}
}

@inproceedings{liu2020robust,
  title={Robust regression for safe exploration in control},
  author={Liu, Anqi and Shi, Guanya and Chung, Soon-Jo and Anandkumar, Anima and Yue, Yisong},
  booktitle={Learning for Dynamics and Control},
  pages={608--619},
  year={2020},
  organization={PMLR}
}

@inproceedings{ahmed2019understanding,
  title={Understanding the impact of entropy on policy optimization},
  author={Ahmed, Zafarali and Le Roux, Nicolas and Norouzi, Mohammad and Schuurmans, Dale},
  booktitle={International conference on machine learning},
  pages={151--160},
  year={2019},
  organization={PMLR}
}

@misc{singh_learning_2018,
    title = {Learning {Stabilizable} {Dynamical} {Systems} via {Control} {Contraction} {Metrics}},
    url = {http://arxiv.org/abs/1808.00113},
    doi = {10.48550/arXiv.1808.00113},
    language = {en},
    urldate = {2025-03-24},
    publisher = {arXiv},
    author = {Singh, Sumeet and Sindhwani, Vikas and Slotine, Jean-Jacques E. and Pavone, Marco},
    month = nov,
    year = {2018},
    note = {arXiv:1808.00113 [cs]},
}

@inproceedings{tang2025deep,
  title={Deep reinforcement learning for robotics: A survey of real-world successes},
  author={Tang, Chen and Abbatematteo, Ben and Hu, Jiaheng and Chandra, Rohan and Mart{\'\i}n-Mart{\'\i}n, Roberto and Stone, Peter},
  booktitle={Proceedings of the AAAI Conference on Artificial Intelligence},
  volume={39},
  number={27},
  pages={28694--28698},
  year={2025}
}

@article{tsukamoto2020robust,
  title={Robust controller design for stochastic nonlinear systems via convex optimization},
  author={Tsukamoto, Hiroyasu and Chung, Soon-Jo},
  journal={IEEE Transactions on Automatic Control},
  volume={66},
  number={10},
  pages={4731--4746},
  year={2020},
  publisher={IEEE}
}

@inproceedings{singh2017robust,
  title={Robust online motion planning via contraction theory and convex optimization},
  author={Singh, Sumeet and Majumdar, Anirudha and Slotine, Jean-Jacques and Pavone, Marco},
  booktitle={2017 IEEE International Conference on Robotics and Automation (ICRA)},
  pages={5883--5890},
  year={2017},
  organization={IEEE}
}

@ARTICLE{tingwei_conformal,
  author={Hsu, Ting-Wei and Tsukamoto, Hiroyasu},
  journal={IEEE Control Systems Letters}, 
  title={Statistical Guarantees in Data-Driven Nonlinear Control: Conformal Robustness for Stability and Safety}, 
  year={2025},
  volume={9},
  number={},
  pages={997-1002},
  doi={10.1109/LCSYS.2025.3578062}}

@INPROCEEDINGS{sihang_conformal,
  author={Wei, Sihang and Ornik, Melkior and Tsukamoto, Hiroyasu},
  booktitle={IEEE Conference on Decision and Control}, 
  title  = {Conformal contraction for robust nonlinear control with distribution-free uncertainty quantification},
  year={2025},
  volume={},
  number={},
  pages={3502-3507},
  doi={10.1109/CDC57313.2025.11312813}}

@inproceedings{tobin2017domain,
  title={Domain randomization for transferring deep neural networks from simulation to the real world},
  author={Tobin, Josh and Fong, Rachel and Ray, Alex and Schneider, Jonas and Zaremba, Wojciech and Abbeel, Pieter},
  booktitle={2017 IEEE/RSJ international conference on intelligent robots and systems (IROS)},
  pages={23--30},
  year={2017},
  organization={IEEE}
}

@article{smirnova2019distributionally,
  title={Distributionally robust reinforcement learning},
  author={Smirnova, Elena and Dohmatob, Elvis and Mary, J{\'e}r{\'e}mie},
  journal={arXiv preprint arXiv:1902.08708},
  year={2019}
}

@article{kalman1960contributions,
  title={Contributions to the theory of optimal control},
  author={Kalman, Rudolf Emil and others},
  journal={Bol. soc. mat. mexicana},
  volume={5},
  number={2},
  pages={102--119},
  year={1960}
}

@article{cimen2012survey,
  title={Survey of state-dependent Riccati equation in nonlinear optimal feedback control synthesis},
  author={Cimen, Tayfun},
  journal={Journal of Guidance, Control, and Dynamics},
  volume={35},
  number={4},
  pages={1025--1047},
  year={2012}
}

@book{royden1988real,
  title={Real analysis},
  author={Royden, Halsey Lawrence and Fitzpatrick, Patrick},
  volume={32},
  year={1988},
  publisher={Macmillan New York}
}

@book{bertsekas1996stochastic,
  title={Stochastic optimal control: the discrete-time case},
  author={Bertsekas, Dimitri and Shreve, Steven E},
  volume={5},
  year={1996},
  publisher={Athena Scientific}
}

@article{jang2025stochastic,
  title={Stochastic Policies, Deterministic Minds: A Calibrated Evaluation Protocol and Diagnostics for Deep Reinforcement Learning.},
  author={Jang, Sooyoung and Yang, Seungho and Choi, Changbeom},
  journal={International Journal of Advanced Computer Science \& Applications},
  volume={16},
  number={10},
  year={2025}
}

@article{montenegro2024learning,
  title={Learning optimal deterministic policies with stochastic policy gradients},
  author={Montenegro, Alessandro and Mussi, Marco and Metelli, Alberto Maria and Papini, Matteo},
  journal={arXiv preprint arXiv:2405.02235},
  year={2024}
}

@misc{zinage2026contractionppocertifiedreinforcementlearning,
      title={ContractionPPO: Certified Reinforcement Learning via Differentiable Contraction Layers}, 
      author={Vrushabh Zinage and Narek Harutyunyan and Eric Verheyden and Fred Y. Hadaegh and Soon-Jo Chung},
      year={2026},
      eprint={2603.19632},
      archivePrefix={arXiv},
      primaryClass={cs.RO},
      url={https://arxiv.org/abs/2603.19632}, 
}

@article{dawson2023safe,
  title={Safe control with learned certificates: A survey of neural lyapunov, barrier, and contraction methods for robotics and control},
  author={Dawson, Charles and Gao, Sicun and Fan, Chuchu},
  journal={IEEE Transactions on Robotics},
  volume={39},
  number={3},
  pages={1749--1767},
  year={2023},
  publisher={IEEE}
}


 





\end{document}